%% file: main_paper.tex
\pdfoutput=1

\newcommand{\CVPR}{CVPR}
\newcommand{\ICCV}{ICCV}
\newcommand{\ECCV}{ECCV}
\newcommand{\ICLR}{ICLR}
\newcommand{\ICML}{ICML}
\newcommand{\NIPS}{NIPS}
\newcommand{\RSS}{RSS}
\newcommand{\ICRA}{ICRA}
\newcommand{\CORL}{CORL}

\newcommand{\conference}{ICLR} %
\newcommand{\paperid}{0000}

\newcommand{\papertype}{ARXIV}

\newcommand{\vlac}{\textsc{VLAC-CUT}}
\newcommand{\TITLE}{HELP: Human-Efficient Large-Scale Robot Post-Training with Rollout Segmentation}

\newcommand{\CoreContributor}{\textsuperscript{*}}
\newcommand{\ProjectLead}{\textsuperscript{\ensuremath{\dagger}}}

\newcommand{\AuthorNames}{%
\makebox[\textwidth][c]{%
\begin{tabular}{c}
Shaopeng Zhai\CoreContributor\ProjectLead,\hspace{0.25em}
Qi Zhang\CoreContributor,\hspace{0.25em}
Tianyi Zhang\CoreContributor,\hspace{0.25em}
Haoran Zhang\CoreContributor,\hspace{0.25em}
Fuxian Huang\CoreContributor \\[0.35em]
Zhanhui Lin,\hspace{0.25em}
Zijun Xu,\hspace{0.25em}
Weinan Zhang
\end{tabular}%
}%
}

\newcommand{\AuthorAffiliation}{%
\makebox[\textwidth][c]{%
\textbf{Continuous Learning Team, Shanghai AI Laboratory}%
}%
}

\newcommand{\AuthorNotes}{%
\makebox[\textwidth][c]{%
{\small
\textsuperscript{*}Core contributors. \quad
\textsuperscript{\ensuremath{\dagger}}Project Lead.
}%
}%
}

\newcommand{\AuthorList}{%
    \AuthorNames \\
    \vspace{-0.6em}\\
    \AuthorAffiliation \\
    \vspace{-0.8em}\\
    \AuthorNotes
}

\documentclass{article}

\usepackage{iclr2024_conference}
\input{math_commands}

\usepackage{times}
\usepackage{graphicx}
\usepackage{amsmath}
\usepackage{amssymb}
\usepackage{booktabs}
\usepackage{multirow}
\usepackage{algorithm}
\usepackage{algpseudocode}
\usepackage{enumitem}
\usepackage{natbib}
\usepackage{url}
\usepackage{xcolor}
\usepackage{tabularx}
\usepackage{array}
\usepackage{hyperref}
\usepackage{fontawesome5}
\usepackage{etoolbox}
\usepackage{wrapfig}
\usepackage{caption}
\usepackage{float}
\usepackage{placeins}
\definecolor{linkcolor}{RGB}{39, 3, 161}
\hypersetup{
  colorlinks=true,
  linkcolor=linkcolor,
  urlcolor=linkcolor
}
\newcommand{\linkicon}[1]{%
  \def\icon{\faLink}%
  \ifstrequal{#1}{huggingface}{\def\icon{\faRobot}}{}%
  \ifstrequal{#1}{hf}{\def\icon{\faRobot}}{}%
  \ifstrequal{#1}{github}{\def\icon{\faGithub}}{}%
  \ifstrequal{#1}{code}{\def\icon{\faGithub}}{}%
  \ifstrequal{#1}{model}{\def\icon{\faCube}}{}%
  \ifstrequal{#1}{homepage}{\def\icon{\faHome}}{}%
  \ifstrequal{#1}{demo}{\def\icon{\faPlayCircle}}{}%
  \ifstrequal{#1}{paper}{\def\icon{\faFile}}{}%
  \ifstrequal{#1}{arxiv}{\def\icon{\faBook}}{}%
  \ifstrequal{#1}{data}{\def\icon{\faDatabase}}{}%
  \icon%
}
\newcommand{\link}[3]{%
  \leavevmode\linkicon{#1}~\href{#3}{\texttt{#2}}%
}
\newcommand{\links}[1]{%
  \par\smallskip\noindent
  \makebox[\linewidth][c]{%
    \begingroup
    \upshape\fontsize{9}{12}\selectfont
    \def\separator{}%
    \renewcommand{\do}[1]{\separator ##1\def\separator{~|~}}%
    \expandafter\docsvlist\expandafter{#1}%
    \endgroup
  }%
  \par
}

\ifx\conference\CVPR
    \title{\TITLE}
\else\ifx\conference\ICCV
    \title{\TITLE}
\else\ifx\conference\ICLR
    \title{\TITLE}
\else\ifx\conference\NIPS
    \title{\TITLE}
\else\ifx\conference\CORL
    \title{\TITLE}
\else\ifx\conference\ICRA
    \title{\LARGE \bf \TITLE}
\fi\fi\fi\fi\fi\fi

\ifx\conference\ECCV
\else\ifx\conference\ICML
\else\ifx\conference\RSS
\else\ifx\conference\ICRA
\else
    \author{\AuthorList}
\fi\fi\fi\fi

\ifx\conference\ICRA
    \author{Albert Author$^{1}$ and Bernard D. Researcher$^{2}$
    \thanks{*This work was not supported by any organization}
    \thanks{$^{1}$Albert Author is with Faculty of Electrical Engineering, Mathematics and Computer Science,
            University of Twente, 7500 AE Enschede, The Netherlands
            {\tt\small albert.author@papercept.net}}%
    \thanks{$^{2}$Bernard D. Researcheris with the Department of Electrical Engineering, Wright State University,
            Dayton, OH 45435, USA
            {\tt\small b.d.researcher@ieee.org}}%
    }
\fi

\ifx\conference\ICLR
    \ifnum\pdfstrcmp{\papertype}{REVIEW}=0
    \else
        \iclrfinalcopy 
\fi\fi

\begin{document}

\ifx\conference\ECCV
    \title{\TITLE}
    \author{First Author\inst{1}\orcidlink{0000-1111-2222-3333} \and
    Second Author\inst{2,3}\orcidlink{1111-2222-3333-4444} \and
    Third Author\inst{3}\orcidlink{2222--3333-4444-5555}}
    
    \authorrunning{F.~Author et al.}
    
    \institute{Princeton University, Princeton NJ 08544, USA \and
    Springer Heidelberg, Tiergartenstr.~17, 69121 Heidelberg, Germany
    \email{lncs@springer.com}\\
    \url{http://www.springer.com/gp/computer-science/lncs} \and
    ABC Institute, Rupert-Karls-University Heidelberg, Heidelberg, Germany\\
    \email{\{abc,lncs\}@uni-heidelberg.de}}
\fi

\ifx\conference\ICML
    \twocolumn[
    \icmltitle{Submission and Formatting Instructions for \\
               International Conference on Machine Learning (ICML 2025)}
    
    \icmlsetsymbol{equal}{*}
    
    \begin{icmlauthorlist}
    \icmlauthor{Firstname1 Lastname1}{equal,yyy}
    \icmlauthor{Firstname2 Lastname2}{equal,yyy,comp}
    \icmlauthor{Firstname3 Lastname3}{comp}
    \icmlauthor{Firstname4 Lastname4}{sch}
    \icmlauthor{Firstname5 Lastname5}{yyy}
    \icmlauthor{Firstname6 Lastname6}{sch,yyy,comp}
    \icmlauthor{Firstname7 Lastname7}{comp}
    \icmlauthor{Firstname8 Lastname8}{sch}
    \icmlauthor{Firstname8 Lastname8}{yyy,comp}
    \end{icmlauthorlist}
    
    \icmlaffiliation{yyy}{Department of XXX, University of YYY, Location, Country}
    \icmlaffiliation{comp}{Company Name, Location, Country}
    \icmlaffiliation{sch}{School of ZZZ, Institute of WWW, Location, Country}
    
    \icmlcorrespondingauthor{Firstname1 Lastname1}{first1.last1@xxx.edu}
    \icmlcorrespondingauthor{Firstname2 Lastname2}{first2.last2@www.uk}
    
    \icmlkeywords{Machine Learning, ICML}
    \vskip 0.3in
    ]
    
    \printAffiliationsAndNotice{Core contributors.} 
\fi

\ifx\conference\RSS
    \title{\TITLE}
    \ifnum\pdfstrcmp{\papertype}{REVIEW}=0
        \author{Author Names Omitted for Anonymous Review. Paper-ID [\paperid]}
    \else
        \author{\authorblockN{Michael Shell}
        \authorblockA{School of Electrical and\\Computer Engineering\\
        Georgia Institute of Technology\\
        Atlanta, Georgia 30332--0250\\
        Email: mshell@ece.gatech.edu}
        \and
        \authorblockN{Homer Simpson}
        \authorblockA{Twentieth Century Fox\\
        Springfield, USA\\
        Email: homer@thesimpsons.com}
        \and
        \authorblockN{James Kirk\\ and Montgomery Scott}
        \authorblockA{Starfleet Academy\\
        San Francisco, California 96678-2391\\
        Telephone: (800) 555--1212\\
        Fax: (888) 555--1212}}
    \fi
\fi

\ifx\conference\ICML
\else
    \maketitle  
\fi

\ifx\conference\ICRA
    \thispagestyle{empty}  
    \pagestyle{empty}    
\fi

\begin{abstract}

\input{sections/abstract}
\end{abstract}

\ifx\conference\RSS
    \IEEEpeerreviewmaketitle  
\fi

\ifx\conference\CORL
\fi


\input{sections/intro}

\input{sections/relatedwork}
\input{sections/framework}
\input{sections/VLAC}
\input{sections/experiements_updated.tex}
\input{sections/experiments_zhr.tex}
\input{sections/contribution.tex}




\clearpage

{
\ifx\conference\CVPR
    \small
\else\ifx\conference\ICCV
    \small
\else\ifx\conference\ICLR
    \small
\else\ifx\conference\NIPS
    \small
\fi\fi\fi\fi
\bibliographystyle{iclr2024_conference}
\bibliography{refs}  
}

\newpage
\appendix
\onecolumn
\input{sections/appendix_vlac}
\input{sections/appendix_teleop}

\end{document}

%% file: math_commands.tex
\usepackage{amsmath,amsfonts,bm}

\def\eqref#1{equation~\ref{#1}}

\def\1{\bm{1}}

\DeclareMathAlphabet{\mathsfit}{\encodingdefault}{\sfdefault}{m}{sl}
\SetMathAlphabet{\mathsfit}{bold}{\encodingdefault}{\sfdefault}{bx}{n}



%% file: sections/abstract.tex
When adapting Vision Language Action (VLA) models to downstream tasks, multiple rounds of post-training are often required to progressively address policy weaknesses. In this report, we focus on maximizing human efficiency during this iterative process, measured by policy improvement and task throughput per unit of human labor and time.

We propose HELP, a Human-Efficient Large-scale robot Post-training pipeline in which two specialized operators supervise twelve robots concurrently. A trained Teleoperator provides high-value remote interventions and recovery demonstrations, while a Floor Operator monitors the robot fleet, triggers takeovers, and performs physical resets. This role specialization improves human efficiency by reducing task switching, lowering operator training costs, and expanding robot interaction coverage. Beyond increasing rollout volume, concurrent supervision also broadens the range of policy behaviors observed by the human team, making recurring failure modes easier to identify and enabling more targeted takeovers, resets, and recovery demonstrations.
To efficiently utilize the large and mixed-quality rollout data, HELP incorporates \vlac, an automatic rollout segmentation critic specifically designed for this setting. It separates autonomous trajectories into progress-making, idle, failure-inducing, and recovery segments. Useful rollout segments are retained and combined with Human-in-the-Loop data for the next post-training round.

Across four real-world manipulation tasks, HELP achieves 80\%--95\% success rates and improves task throughput by 1.7$\times$--4.2$\times$ over the base model. Under matched HITL recovery budgets, \vlac further amplifies throughput gains by 1.20$\times$--3.43$\times$ and success-rate gains by 1.50$\times$--3.00$\times$ over HITL-only updates.

\links{
  \link{github}{Code:VLAC-CUT}{https://github.com/InternRobotics/VLAC-cut},
  \link{huggingface}{Model:VLAC-CUT}{https://huggingface.co/InternRobotics/VLAC-Cut},
  \link{huggingface}{Benchmark:VPB}{https://huggingface.co/datasets/InternRobotics/VLAC-Cut-Benchmark},
  \link{huggingface}{Dataset}{https://huggingface.co/datasets/InternRobotics/VLAC-Cut-FullData},
}

%% file: sections/intro.tex
\section{Introduction}
After pre-training, generalist policies like Vision Language Action (VLA) models require post-training to achieve reliable performance on downstream tasks. This phase typically begins with collecting task-specific data to fine-tune the VLA. However, because human operators cannot foresee all potential edge cases during the initial data collection, the models fine-tuned on this preliminary data inevitably exhibit flaws. Consequently, post-training inherently involves multiple iterations. In each round, the policy is evaluated, and its specific failure modes are explicitly recorded to guide targeted data collection in the subsequent phase.
This iterative process is crucial for progressively mitigating weaknesses and improving the actual task success rate. However, each iteration also requires humans to evaluate failures, intervene in difficult states, reset physical scenes, and collect or curate new training data. As these costs accumulate across repeated post-training rounds, human labor becomes a central bottleneck for scaling real-world VLA adaptation. Therefore, the key question is not only how to improve the policy, but how much policy improvement and task throughput can be obtained per unit of human labor and time.

Currently, real-world post-training paradigms generally fall into three categories. The first is real-world Reinforcement Learning (RL). This encompasses both fully online methods, where the boundaries between training iterations are entirely blurred, and near-online methods with clearly defined iterations, such as the approach used in pi0.6. These approaches require the model to autonomously collect data in the real world, generating a mixture of successful and failed rollouts. Unlike traditional supervised learning that relies purely on positive samples, RL methods can effectively extract useful training signals from these mixed trajectories. The second paradigm focuses on targeted human data collection, for example, \citep{intelligence2025pi06vla}. Here, human operators conduct specific data collection to address the policy weaknesses identified in the previous round. This process yields a substantial volume of recovery data, helping the policy learn to recover from common errors. The third and most prevalent approach integrates the former two. It begins with an offline dataset, and as the model generates autonomous real-world rollouts, human operators can intervene at any moment to provide Human-in-the-Loop (HITL) recovery data. These diverse data sources are subsequently combined, allowing the policy to be trained through a hybrid strategy of RL and Supervised Fine Tuning (SFT).

However, constrained by the limited exploration capabilities of current pre-trained policies, VLAs often struggle to autonomously execute correct manipulation behaviors when facing unfamiliar corner cases. Consequently, Human-in-the-Loop (HITL) intervention remains an absolute necessity. While some existing works attempt to learn a better policy from limited static datasets, our objective is fundamentally different. We do not aim to optimize sample efficiency on fixed data. Instead, we focus on maximizing how effectively limited human resources can be deployed to actively collect data that directly addresses policy weaknesses and resolves task bottlenecks. We define human efficiency as the post-training policy performance gain, specifically in success rate and system throughput, achieved per unit of human labor and time. Therefore, the core challenge lies in optimizing human roles and pairing them with an appropriate system framework and a dedicated generalist model. By lowering operator training costs and ensuring singular task assignments, this model-assisted paradigm enables humans to focus exclusively on the most critical tasks and the most urgently needed data.

To address this challenge, we introduce HELP, a human-efficient pipeline for large-scale robot post-training. HELP combines role-specialized multi-robot supervision, targeted human takeover, autonomous rollout collection, and process-level rollout segmentation into a unified iterative post-training framework. Beyond increasing rollout volume, concurrent supervision also broadens the range of policy behaviors observed by the human team, making recurring failure modes easier to identify and enabling more targeted takeovers, resets, and recovery demonstrations. A key component of HELP is \vlac, a Vision-Language-Action Critic whose ``cut'' operation denotes its role in segmenting autonomous robot trajectories (see Fig.~\ref{fig:vlac_cut}). \vlac\ evaluates task-conditioned rollouts at the process level and separates reusable progress or recovery segments from idle and failure-inducing portions. It is designed specifically for HELP, serving as the rollout segmentation module that connects autonomous data collection, HITL recovery data, and subsequent policy training. HELP first improves human efficiency through role-specialized multi-robot supervision, while \vlac\ further amplifies the learning return from the resulting autonomous rollout data. In the real-world experiments, we further quantify the contribution of \vlac\ to supervision utilization using a human-supervision amplification factor under matched HITL recovery budgets.

Our contributions are threefold:
\begin{itemize}[leftmargin=*]
\item \textbf{Human-efficiency-centered post-training formulation:} We identify human labor, rather than only robot interaction count or offline sample efficiency, as a central bottleneck in real-world VLA post-training. We therefore formulate human efficiency as the policy improvement, measured by success rate and task throughput, obtained per unit of human labor and time.

\item \textbf{HELP: role-specialized large-scale robot post-training.} We introduce HELP, a distributed multi-robot post-training pipeline in which two operators, assigned to two specialized roles, supervise twelve physical robots concurrently. By separating high-skill teleoperation from on-site monitoring and reset, HELP reduces task switching, lowers operator training barriers, and concentrates human effort on high-value takeover and recovery data. Beyond increasing interaction throughput, concurrent fleet supervision broadens policy-behavior coverage, helping the Floor Operator identify recurring failure modes and collect targeted corrective data.

\item \textbf{\vlac-based rollout segmentation in HELP.} We introduce \vlac\ as a process-level multimodal rollout segmentation critic within HELP. \vlac\ segments autonomous rollouts into progress-making, failure-inducing, idle, and recovery segments, allowing HELP to preserve useful robot-collected data while filtering behaviors that would otherwise reinforce inefficient trial-and-error patterns. Under matched HITL recovery budgets, this rollout curation further amplifies the throughput and success-rate gains obtained from the same human-supervision budget.
\end{itemize}
We validate HELP in real-world multi-robot post-training experiments using a two-operator, twelve-robot deployment. Controlled comparisons under matched HITL recovery budgets further show that \vlac\ amplifies throughput gains by 1.20$\times$--3.43$\times$ and success-rate gains by 1.50$\times$--3.00$\times$ relative to HITL-only updates. Overall, our work shifts the focus of robot post-training from raw sample efficiency to systemic human efficiency, establishing a practical framework for high-throughput, human-robot collaborative learning at scale.

%% file: sections/relatedwork.tex
\section{Related work}
\label{sec:related}

\hspace{1.5em}\textbf{Real-world post-training for robot policies} has increasingly moved from static imitation learning toward iterative interaction, where autonomous rollouts expose policy weaknesses and subsequent data collection or optimization improves downstream performance. Real-world RL has demonstrated strong potential in locomotion\citep{smith2024grow,smith2023demonstrating} and dexterous or general manipulation\citep{pmlr-v229-hu23a}, but deploying such methods on physical robots remains expensive because each trial consumes real time, hardware lifetime, and human supervision. Many online or near-online systems therefore reuse previously collected trajectories through self-imitation, hindsight relabeling, or offline-to-online updates\citep{kumar2024practice,zhou2024autonomous,luo2024hilserl}. For smaller manipulation policies, human-in-the-loop pipelines commonly begin with a small set of expert demonstrations and then combine robot interaction with human corrections to improve sample efficiency\citep{kang2025forget,zhou2024efficient,li2025reinforcement}. Recent VLA-oriented post-training methods further benefit from pretrained multimodal priors, which increase the chance of discovering useful behaviors and make it possible to learn from mixed successful, failed, and recovered rollouts\citep{chen2025conrft,lv2025flow,park2025flow}. This trend has accelerated over the past year: $\pi^{*}_{0.6}$ studies how VLAs can improve from real-world deployment data that combine demonstrations, on-policy rollouts, and expert teleoperated interventions\citep{physicalintelligence2025pi06}; ROVE targets humanoid VLA post-training from imperfect human interventions and prioritizes high-value behaviors within mixed-quality trajectories\citep{xiao2026rove}; SOP and Learning While Deploying scale online post-training to robot fleets that stream on-policy experience and human intervention signals to a centralized learner\citep{pan2026sop,wang2026learningwhiledeploying}. Complementary work reduces the cost of physical interaction through world-model or sim-real post-training, using virtual environments, simulated RL, or latent world-model scoring to improve VLAs under limited real-world data\citep{xiao2025worldenv,shi2026beyond,sun2026atomvla}.

Despite this progress, real-world VLA post-training still depends critically on how raw interaction data are converted into reliable supervision. VLA policies differ substantially in their action interfaces: some produce discrete action tokens or structured textual actions\citep{kim24openvla,pertsch2025fast,zhai2024buildingopenendedembodiedagent}, whereas others output continuous actions through diffusion or flow-based decoders\citep{black2024pi_0,chi2023diffusion,kim2025fine}. This heterogeneity makes it difficult to apply a single token-centric RL recipe across models, and existing methods often resort to value-guided sample selection, behavior-cloning regularization, conservative value learning, or trajectory filtering to stabilize policy improvement \citep{he2024aligniql,lv2025flow,park2025flow,dingconsistency,physicalintelligence2025pi06,wang2026learningwhiledeploying}. However, these strategies usually treat each rollout as a full episode or rely on coarse success/failure labels, while real robot trajectories often contain standard execution, failure-inducing detours, repeated trial-and-error, and useful recovery behaviors within the same episode. In this setting, blindly training on unedited rollouts can teach the policy to imitate avoidable failures, whereas discarding entire imperfect trajectories wastes valuable recovery data. \vlac\ addresses this rollout segmentation bottleneck directly. It performs process-level rollout segmentation, removes failure-inducing portions, and preserves both correct execution and recovery segments. This rollout-segmentation formulation is especially important for high-throughput multi-robot post-training, where manual inspection does not scale and human effort should be concentrated on high-value intervention rather than offline data cleaning.

\paragraph{Learned reward and progress models for embodied AI.}
A central challenge in embodied learning is replacing hand-crafted task rewards with scalable reward, value, or progress estimators defined over vision and language. Early approaches derived proxy measures of task advancement from human videos, trajectories, or goal-conditioned visual embeddings (e.g., VIP, LIV, R3M) \citep{sermanet2016unsupervised, shao2020concept, chen2021learning, ma2022vip, ma2023liv, nair2022r3m, yang2023robot, sontakke2024roboclip}, but these are often task-specific and lack fine-grained language grounding. Subsequent foundation-model methods sought to shape or synthesize rewards directly from language and vision \citep{ma2023eureka, rocamonde2023vision, cui2025grove, lin2024navigating}, framing reward estimation as binary success prediction, image comparison, temporal reordering, or relative progress scoring \citep{du2023vision, wang2024rlvlmf, venkataraman2024offlinevlm, luu2025erlvlm, singh2025varp, alakuijala2024videolanguagecritic, ma2024vision, chen2025sarm}. While substantially improving scalability, these formulations highlight a persistent trade-off: success classifiers are often too coarse for dense shaping, and embedding distances struggle to capture complex intermediate progress \citep{du2023vision, lynch2020learning, andrychowicz2017hindsight, escontrela2023video, lee2021generalizable, ma2024vision}. To address this gap, recent work has shifted toward general-purpose, step-aware process reward modeling (PRM) to provide dense progress supervision. Methods such as VLAC, Robo-Dopamine, and Rewarding DINO extract dense progress deltas and step-aware signals to overcome the limitations of sparse or uniform reward designs \citep{zhai2025vlac, tan2025robodopamine, krack2026rewarding}. Concurrently, VLM-based estimators like RoboReward, ProgressLM, and Large Reward Models explicitly frame progress tracking as a long-horizon reasoning or sequence modeling problem, providing discretized progress labels and contrastive signals for closed-loop refinement \citep{lee2026roboreward, zhang2026progresslmprogressreasoningvisionlanguage, wu2026large}. Further extending these paradigms, Robometer incorporates inter-trajectory preference learning to resolve ambiguities in suboptimal demonstrations \citep{liang2026robometer}, complementing a broader effort to leverage step decomposition, distance-to-goal estimation, and binary completion queries as robust interfaces for scalable reward supervision \citep{chen2025sarm, ghasemipour2025self, yu2026chi_, mao2026arm, intelligence2025pi06vla, chen2026topreward}.

\paragraph{Benchmarks for robot progress evaluations.}
Reward models have become increasingly important in modern foundation-model training, especially in post-training and reinforcement learning for large language models \citep{shao2024deepseekmathpushinglimitsmathematical, deepseekai2025deepseekr1incentivizingreasoningcapability}. Accordingly, benchmarks such as RewardBench, RewardBench~2, VLRewardBench, and Multimodal RewardBench have been introduced to evaluate reward models in language and general multimodal settings \citep{lambert2024rewardbench, malik2025rewardbench2, li2024vlrewardbench, yasunaga2025mmrewardbench}. In embodied settings, however, benchmark construction remains limited and fragmented. OpenGVL evaluates temporal progress prediction from shuffled trajectory frames using a Value-Order Correlation metric, with a particular emphasis on automated data curation and filtering \citep{budzianowski2025opengvl}. ManiRewardBench advances reward benchmarking on real-world manipulation with subtask-level temporal annotations, enabling evaluation of progress sensitivity, completion detection, and cross-embodiment robustness \citep{chen2026topreward}. RoboRewardBench broadens reward evaluation to diverse real-robot tasks and explicitly includes unsuccessful and near-miss trajectories, but is primarily framed around short-horizon, end-of-episode coarse progress rewards \citep{lee2026roboreward}. Robometer further introduces held-out evaluation sets that probe reward alignment, trajectory ranking, and success--failure separation across unseen institutions, embodiments, camera viewpoints, and scenes \citep{liang2026robometer}. Adjacent to reward modeling, RoboFAC provides a failure-centric benchmark for task understanding, failure diagnosis, and hierarchical correction on simulated and real-world robot videos \citep{ye2025robofac}. Overall, existing embodied benchmarks capture complementary aspects of temporal progress, reward calibration, and failure understanding, yet still leave room for evaluating long-horizon, process-level reward and value estimation with richer temporal structure and finer-grained supervision on realistic robotic trajectories.

%% file: sections/framework.tex
\section{The HELP Framework}
\label{sec:framework}

\subsection{HELP System Architecture}
\label{subsec:system_architecture}

HELP is designed to maximize human efficiency in large-scale real-world robot post-training, where a small number of operators must supervise a fleet of concurrently operating robots. Its system architecture is explicitly designed around the operational demands and collaborative workflows of the two specialized human roles.

\begin{figure}[htbp]
    \centering
    \includegraphics[width=\linewidth]{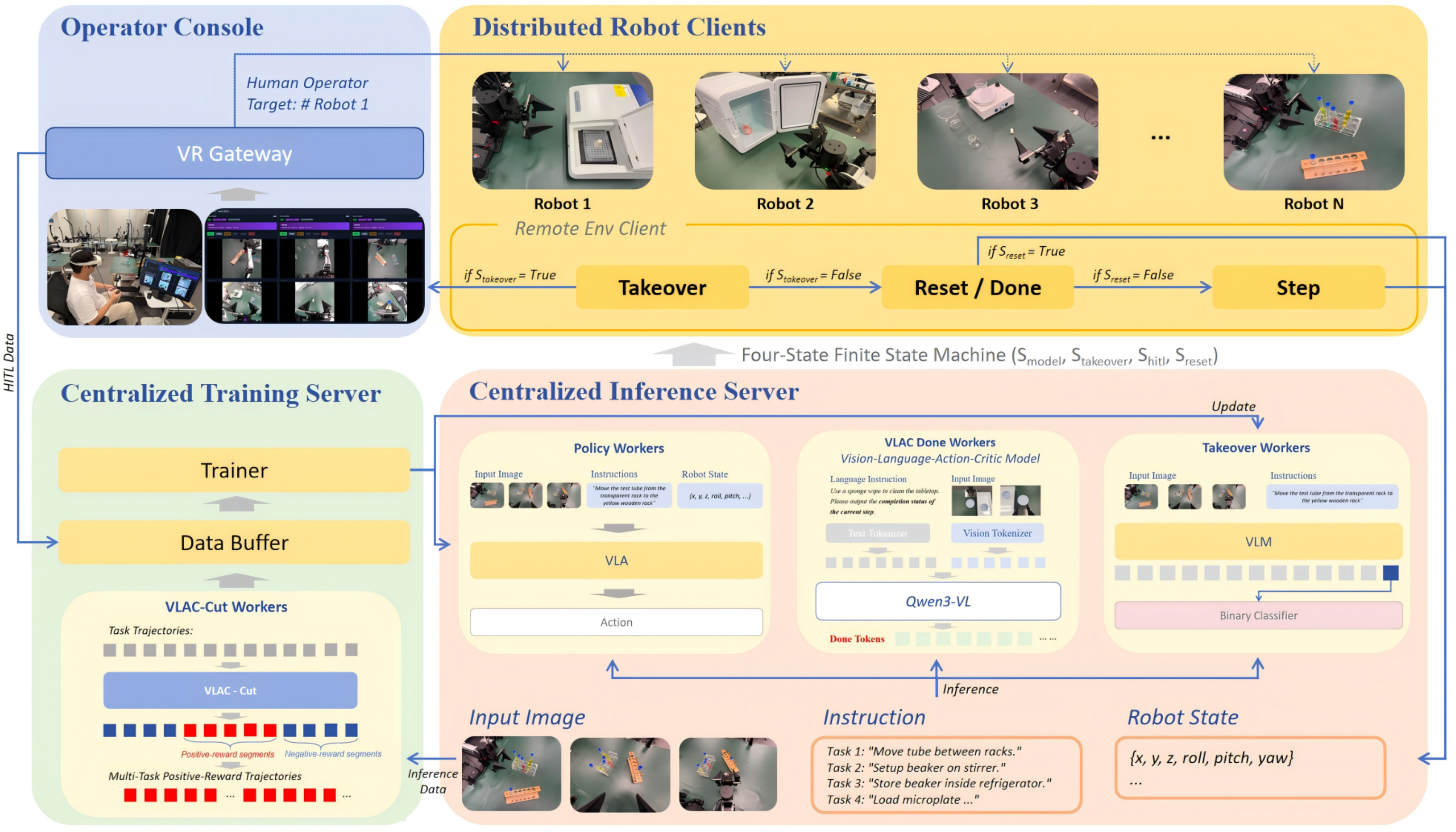}
    \caption{System architecture of HELP, our distributed human-efficient robot post-training framework.}
    \label{fig:system_architecture}
\end{figure}

First, the Teleoperator undergoes extensive specialized training and incurs high training costs. To maximize their productivity, HELP must allow them to concentrate exclusively on high quality data collection without wasting time moving around or physically resetting scenes. This necessitates a seamless and stationary remote control interface. Second, the Floor Operator simultaneously manages multiple physical robots. Because they spend minimal time at any single robotic station while constantly observing the policy in action, they develop a profound understanding of the model capabilities and specific failure cases. Observing concurrent executions across the robot fleet not only increases data throughput, but also makes recurring policy weaknesses more visible and allows subsequent interventions to focus on the most informative failure states. Their cognitive effort must be concentrated on issuing takeover signals and manually resetting scenes to setups where the policy frequently fails. Consequently, the architecture must provide dedicated and lightweight communication modules for the Floor Operator to instantly broadcast reset and takeover commands. Finally, to sustain this continuous collaboration, HELP must seamlessly interleave Human-in-the-Loop (HITL) data with autonomous robot rollouts to support both online and offline training. This holistic pipeline is heavily supported by VLMs tasked with processing trajectory data, assessing task completion, and proactively predicting the necessity of takeovers. Driven by these human-centric requirements, we logically partition our distributed software architecture into three primary layers, as illustrated in Fig.~\ref{fig:system_architecture}.

\begin{itemize}[leftmargin=*]
    \item \textbf{Centralized GPU workers:} Deployed on a centralized server, these GPU workers manage parallel inference and continuous policy optimization using the Ray distributed framework. At each control step, an asynchronous coordination module aggregates multi modal observations from the distributed robotic fleet and routes them to parallel inference workers through dynamic load balancing. These workers execute the concurrent inference of the VLA policy $\pi_\theta$, a predictive takeover model $\phi$, and a task termination model $\psi$:
    \begin{equation}
    \mathbf{a}_t^i = \pi_\theta(s_t^i), \quad \gamma_t^i = \phi(s_t^i), \quad d_t^i = \psi(s_t^i)
    \label{eq:joint_inference_fixed}
    \end{equation}
    where $\mathbf{a}_t^i$ is the action chunk, $\gamma_t^i \in \{0, 1\}$ indicates the model prediction on whether a takeover is currently required at step $t$, and $d_t^i \in \{0, 1\}$ represents the model estimated task completion state. Following this parallel computation, the server bundles these outputs and directly transmits them back to the respective robotic client. The client then utilizes these flags to determine its execution state, where a takeover signal ($\gamma_t^i=1$ or a manual override) or a reset signal ($d_t^i=1$) will preempt the standard action execution. Once an episode concludes, the autonomous rollout trajectories are segmented by the learned \vlac{} model. The resulting curated rollout segments are then integrated into a centralized data service managed by a dedicated Ray worker. Concurrently, an asynchronous training pipeline samples from this data service to update $\theta$ and $\phi$ via gradient descent, asynchronously broadcasting the updated parameters to the inference workers to eliminate training induced latency.

    \item \textbf{Distributed Robot Clients:} Running on each onboard computer, the Robot Clients interface directly with the robot and manage real time data routing. On one hand, the client continuously captures robot proprioceptive data and camera images, streaming them to the Centralized GPU Workers for parallel inference, while simultaneously executing the predicted action chunks returned by the server. On the other hand, it serves as the direct communication endpoint, receiving and executing intervention signals routed through the server.

    To enable a flexible and timely takeover experience for the human operators, the client utilizes a dual process and asynchronous multi threaded design. This architecture explicitly isolates the hardware control loop from the overhead of network I/O and VR telemetry streams. By leveraging ZeroMQ for robust asynchronous command fetching and local inter process communication to manage incoming payloads, the system structurally decouples communication from robot execution.

    Driven by these incoming streams, the client finite state machine transitions dynamically via a priority hierarchy:
    \begin{equation}
    \mathcal{S}_{\text{client}}^i \leftarrow 
    \begin{cases}
    \mathcal{S}_{\text{takeover}} & \text{if } u_t^i = \operatorname{takeover}, \\
    \mathcal{S}_{\text{hitl}} & \text{if } \mathcal{S}_{\text{client}}^i = \mathcal{S}_{\text{takeover}} \text{ and VR streaming is active}, \\
    \mathcal{S}_{\text{reset}} & \text{if } u_t^i = \operatorname{reset}, \\
    \mathcal{S}_{\text{model}} & \text{if } u_t^i = \operatorname{step}(\mathbf{a}_t^i).
    \end{cases}
    \label{eq:fsm_mapping}
    \end{equation}
    In the autonomous execution state $\mathcal{S}_{\text{model}}$, the client unrolls the received action chunks. When a takeover signal arrives from the Centralized GPU Workers, the client instantly purges stale actions and transitions through the standby state $\mathcal{S}_{\text{takeover}}$ into the active human control state $\mathcal{S}_{\text{hitl}}$ as soon as the VR stream synchronizes, guaranteeing a rapid and unimpeded handover for the Teleoperator. If a reset signal is issued, the client immediately aborts execution and notifies the Centralized GPU Workers to terminate the current episode recording, prompting the server to save and process the trajectory data. The client then enters the suspension state $\mathcal{S}_{\text{reset}}$, securely locking the robot to allow the Floor Operator to safely perform the scene reset.

    \item \textbf{Operator Console:} Serving as the unified human interface, the Operator Console consists of a web frontend, a VR application, a hardware interface for the Floor Operator, and a centralized backend service. The Teleoperator utilizes the web frontend to monitor real time visual feeds from individual robots and to explicitly dictate the routing destination for the VR action commands. Once a target is assigned, the Teleoperator directly controls the robot to execute manipulation tasks through the VR application. To improve intervention efficiency under transient VR command latency or loss, the console also incorporates a lightweight VLA-based teleoperation assistance module, detailed in Appendix~\ref{app:teleop_assistance}. Concurrently, the interface for the Floor Operator pairs with a portable Bluetooth keyboard, enabling them to rapidly force any specific robot into $\mathcal{S}_{\text{takeover}}$ while monitoring the active robots on site. Underpinning these interactive modules is the backend service which manages the critical communication middleware. For VR control, it employs a dynamic UDP routing gateway to seamlessly redirect high frequency telemetry streams to the designated robot. For event handling, it captures the Bluetooth keyboard inputs and leverages ZeroMQ sockets to instantly trigger $\mathcal{S}_{\text{takeover}}$ for the corresponding robot.
\end{itemize}

\subsection{Role-Specialized Supervision in HELP}
\label{subsec:pipeline}
In HELP, the human supervision workload is divided into two specialized roles: the Teleoperator and the Floor Operator. In our typical operational configuration, this division allows two personnel to supervise twelve robots concurrently. By ensuring that each person focuses on a designated responsibility, HELP reduces the cognitive overhead and efficiency degradation typically caused by task switching. Furthermore, isolating these duties simplifies the onboarding process, enabling targeted, specialized training for each individual role and making skilled teleoperation labor available for high-value takeover and recovery data. This multi-robot supervision setup therefore serves not only as a mechanism for scaling data collection, but also as an active failure-mode discovery process that guides targeted takeovers, resets, and recovery demonstrations.

\begin{itemize}[leftmargin=*]
\item \textbf{Teleoperator:}
The Teleoperator dedicates their entire cognitive focus exclusively to remote manipulation. A proficient data collector must exhibit both optimal behavioral patterns and highly consistent operational habits, as these factors directly determine the quality of the collected data, which in turn dictates the efficiency of post-training. For instance, in a test tube insertion task, the operator must deliberately reduce the manipulation speed during the high-precision insertion phase and appropriately elevate the robot end effector to increase the clearance between the tube base and the target hole. Training a skilled Teleoperator to internalize these specific manipulation patterns often requires weeks or even months. By keeping the Teleoperator responsibilities strictly singular, we can concentrate our training efforts on these essential skills while significantly lowering their overall cognitive workload. During operation, the Teleoperator remains stationary and uses a VR interface to monitor real-time camera feeds. When a robot is identified for a takeover by either the Floor Operator or the predictive model, a notification is transmitted to the Teleoperator. The system then seamlessly routes the corresponding camera feed to the VR interface, enabling the operator to quickly transmit action commands back to the robot. This remote operational paradigm reduces physical locomotion and time wasted walking between different robots. By isolating this role from manual scene resetting and physical movement, the Teleoperator achieves high data collection efficiency with fewer physical and cognitive interruptions.
\item \textbf{Floor Operator:}
The Floor Operator is responsible for executing resets and continuously monitoring the policy execution on site, selectively forcing specific robots into $\mathcal{S}_{\text{takeover}}$. Because they observe the autonomous rollouts over extended periods, the Floor Operator develops a highly intuitive and profound understanding of the policy behavior. They know exactly when the model is likely to fail, making them the ideal judge for determining the precise moment for a robot to enter the takeover state. Furthermore, their deep awareness of the policy weaknesses allows them to manually reset the environment to states where the policy frequently fails, thereby maximizing the collection of high-value data in the exact scenarios that expose model limitations. In practice, their operational duties remain tightly focused. They actively monitor the active robots and use a portable Bluetooth keyboard to instantly force a specific robot into $\mathcal{S}_{\text{takeover}}$. When a robot halts and awaits a reset, they perform these targeted resets. By keeping their responsibilities strictly singular, the system empowers the Floor Operator to fully leverage their cognitive understanding of the policy behavioral characteristics. As the individual who most comprehensively observes the model in action, they play a critical role in creating the conditions to collect targeted data that addresses the policy shortcomings.
\end{itemize}

\begin{figure}[!t]
    \centering
    \includegraphics[width=\linewidth]{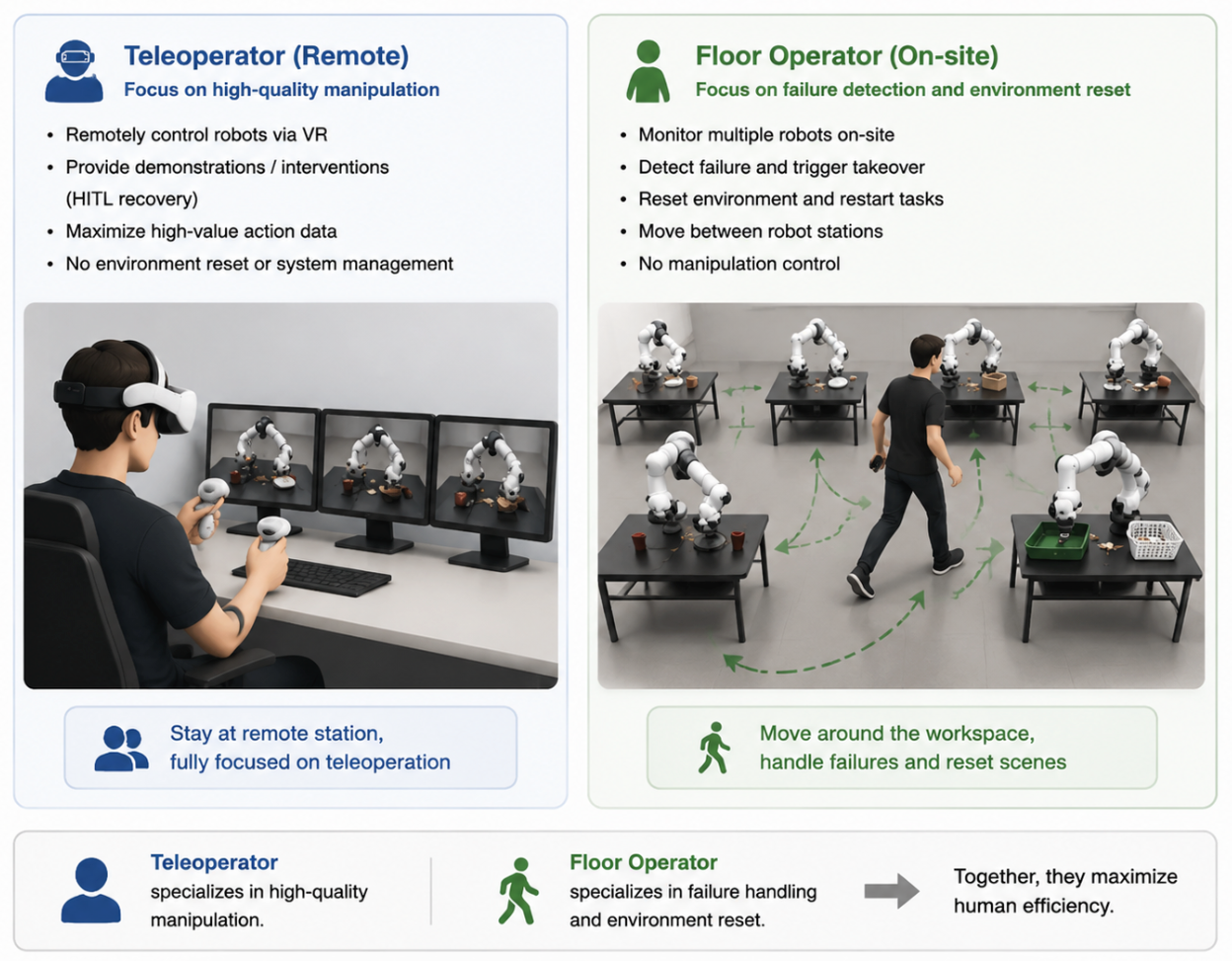}
    \caption{Role specialization between the Teleoperator and the Floor Operator in HELP.}
    \label{fig:characters}
\end{figure}

\subsection{Post-Training in HELP}
\subsubsection{Data Processing by \vlac{}}
During HELP data collection, the system accumulates two distinct categories of interaction data. The first is rollout data generated entirely by the VLA policy. The second is Human-in-the-Loop data, which is collected when either the predictive takeover model or the Floor Operator forces the robot into $\mathcal{S}_{\text{takeover}}$, with the subsequent actions executed by the Teleoperator. Consequently, a complete trajectory within HELP exhibits one of two structural configurations: a homogeneous trajectory composed exclusively of VLA rollouts, or a hybrid trajectory consisting of an initial VLA rollout prefix followed by a Human-in-the-Loop recovery suffix.
In addition to serving as recovery demonstrations for policy post-training, the manually triggered HITL events also provide supervision for the takeover prediction model. Specifically, the observation at which the Floor Operator issues a takeover command is labeled as a positive takeover example, while observations sampled from normal autonomous rollouts are used as negative examples to train a 2B Qwen-VL based takeover predictor.

As the policy is continuously trained on the accumulated recovery data, it gradually develops the capability to recover from failure states. Consequently, the raw VLA rollouts generated during continuous data collection frequently contain repetitive trial and error, where the policy continuously generates errors and then recovers from them. While this continuous trial and error maintains a high task success rate, directly incorporating these unedited trajectories into the training set would teach the policy to replicate these repetitive actions, ultimately reducing its overall task execution throughput.

To address this issue, we employ \vlac\ to segment autonomous rollout trajectories into four types: progress-making and recovery segments, which are retained for policy training, and idle and failure-inducing segments, which are excluded from policy optimization. Our VLA is trained with a flow-matching objective, and in the setting considered in this work, we did not find a sufficiently reliable way to directly use failure-inducing action segments as negative learning signals under this objective. Training on unsegmented rollouts would therefore reinforce not only useful behaviors, but also the trial-and-error actions that caused the failure. We therefore combine the retained progress-making and recovery segments into the curated rollout dataset, which is subsequently mixed with HITL recovery data for the next optimization cycle.

\subsubsection{Algorithm}
Training a unified policy on data from a distributed multi-robot fleet introduces inherent data imbalance and distribution shift challenges. First, a distribution mismatch exists between the standard rollout data and the Human-in-the-Loop recovery data, because the action distribution required to correct a failure differs significantly from that of standard execution. Second, the physical heterogeneity across different environments, including variations in object layouts, camera viewpoints, and task definitions, induces cross-environment interference. Consequently, optimizing the policy exclusively on newly collected continuous streams often degrades its performance on other environments and causes catastrophic forgetting of previously learned capabilities. To mitigate these distribution shifts and manage the continuous learning process, we formalize two distinct optimization strategies across training cycles:
\begin{itemize}[leftmargin=*]
\item \textbf{Fully Online Incremental Optimization:}
As data collection proceeds continuously, the server initiates an immediate incremental training update once the accumulated data reaches a specified threshold. To mitigate catastrophic forgetting and cross environment interference inherent to this streaming continual learning setup, we integrate the ConSFT algorithm \citep{zhang2026preserving}. This approach establishes a self regulating learning dynamic that scales the optimization gradients based on per sample model confidence, eliminating the need for parallel reference networks or historical data buffers. The optimization objective is formulated as:
\begin{equation}
\mathcal{J}_{\mathrm{ConSFT}}(\theta) = \mathrm{sg}\left[ \exp\left(-\frac{\mathcal{L}_{\mathrm{SFT}}(\theta)}{\tau}\right) \right] \cdot \mathcal{L}_{\mathrm{SFT}}(\theta)
\label{eq:consft_objective}
\end{equation}
where $\tau > 0$ is a temperature parameter regulating the conservative scaling sensitivity, and $\mathrm{sg}[\cdot]$ denotes the stop gradient operator. During the initial phases of incremental adaptation, out of distribution behaviors, such as some recovery actions, generate large losses. This objective exponentially suppresses their gradient contributions to prevent disruptive parameter updates. As the policy gradually masters these novel behaviors, its confidence increases, causing the associated loss to decrease and consequently amplifying the update weights. This continuous feedback loop ensures a progressive and stable learning of new out of distribution behaviors while preserving previously learned capabilities.

\item \textbf{Periodic Batched Optimization:}
This configuration operates with discrete training intervals. Following each policy update, the system accumulates a new batch of interaction trajectories. Once a predefined dataset size is reached, this current batch is combined with the entire historical dataset from all preceding rounds for joint training. By aggregating and mixing historical data across all robots and task configurations, this approach resolves distribution shifts and mitigates catastrophic forgetting, enabling standard supervised fine tuning via flow matching.
\end{itemize}

%% file: sections/VLAC.tex
\section{\vlac}
\label{sec:vlac_cut}

\begin{figure}[t]
    \centering
    \includegraphics[width=0.95\linewidth]{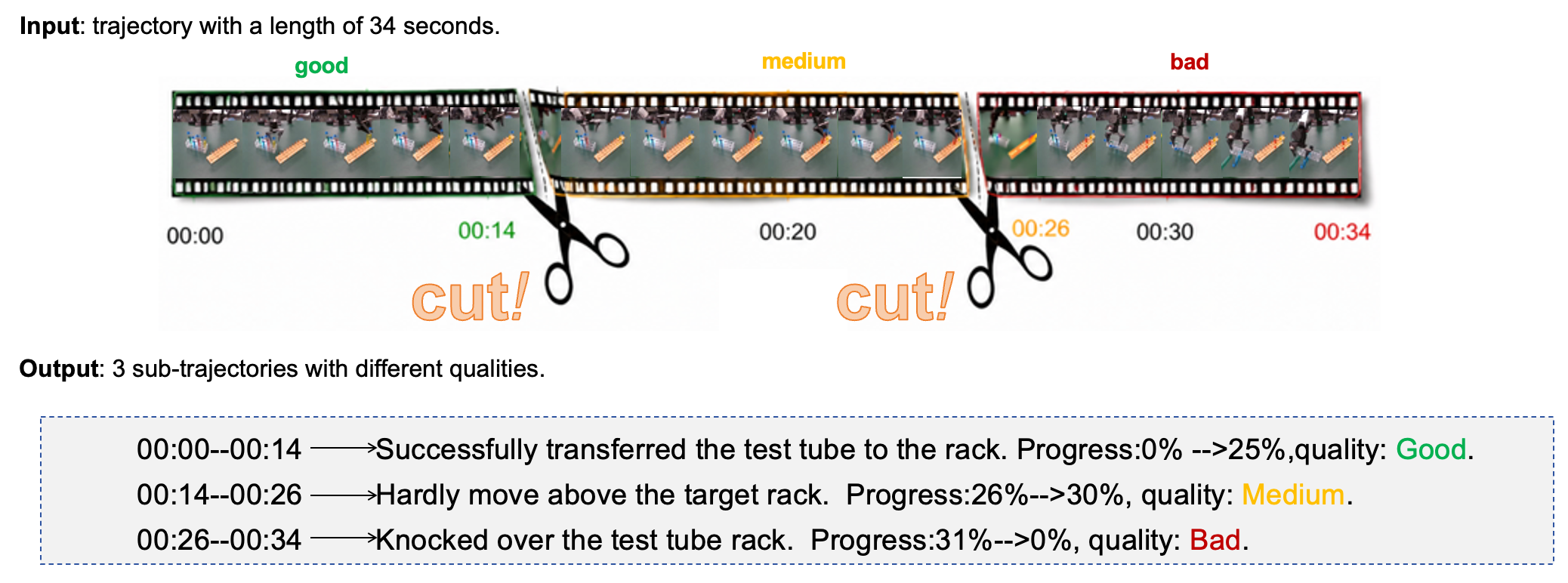}
    \caption{Overview of the \vlac\ rollout segmentation pipeline. Given a task-conditioned autonomous rollout trajectory, \vlac\ estimates process-level progress, diagnoses local state changes, and segments the trajectory into training-useful and training-harmful portions.}
    \label{fig:vlac_cut}
\end{figure}

\subsection{\vlac{} as a Rollout Segmentation Critic}
\label{sec:vlac_cut_critic}

HELP relies on large-scale autonomous rollout segmentation to amplify limited human supervision.
However, a raw autonomous rollout trajectory is rarely uniformly useful for policy post-training.
It may contain correct progress, idle motion, failure-inducing detours, repeated trial-and-error, and recovery behavior within the same episode.
Training on the entire trajectory can therefore teach the policy to imitate avoidable failures, while discarding the entire trajectory wastes useful robot-collected progress and recovery data.

As shown in Fig.~\ref{fig:vlac_cut}, \vlac\ is designed to turn raw autonomous rollouts into training-ready segments.
Given a task instruction and a rollout trajectory, prefix, or candidate segment, \vlac\ estimates signed task progress, diagnoses the visual and action-level causes of local progress changes, and predicts cut points that separate segments with different training value.
Its output tells the post-training data service which parts of a rollout should be kept, which parts should be removed, and which recovery portions should be mixed with HITL data for the next policy update.
Unlike a general-purpose reward function, which typically scores a state from its instantaneous state representation, our critic evaluates the state in the context of the trajectory in which it appears.
As a result, the same state may receive different scores in different trajectories, depending on how the policy reached that state and what consequences its subsequent actions are expected to produce.
In this sense, \vlac\ is closer to a policy-conditioned value estimator than a universal reward function, because its score reflects the expected downstream task progress under the deployed policy, conditioned on the current state and trajectory context, rather than an immediate, policy-agnostic reward assigned to the state alone.

We use four segment types throughout the paper: \textit{progress-making} segments correspond to standard execution that should reinforce correct behavior; \textit{idle} segments contain little task-relevant progress; \textit{failure-inducing} segments move the task state away from the goal or create avoidable errors; and \textit{recovery} segments bring the system back from an error state toward task completion.
The post-training data service keeps progress-making and recovery segments, removes failure-inducing and idle portions, and combines the curated rollout segments with HITL recovery data for the next optimization cycle.

This formulation differs from endpoint success filtering.
End-state labels can identify whether an episode eventually succeeds, but they cannot distinguish an efficient execution from a trial-and-error success, nor can they identify which intermediate actions caused stagnation, regression, or recovery.
\vlac\ instead treats task progress as a temporal process variable, making autonomous rollout segmentation compatible with the human-efficiency objective of HELP.

\subsection{Process Supervision for \vlac}
\label{sec:progress_annotation_dataset}

Training such a critic requires supervision beyond demonstrations and final success labels.
We therefore construct a Progress Annotation Dataset in which each annotated episode contains a task instruction, a task-level plan, sparse timestamped progress points, diagnostic language, and optional grounding/action annotations.
The progress labels are signed: positive values indicate movement toward task completion, zero denotes the initial state, and negative values represent regressive states worse than the initial setup.
This schema directly supervises the distinctions needed by \vlac: partial progress, stagnation, regression, failure, and recovery.

The dataset combines public robot sources with an in-house real-world ARX-data collection.
Public datasets provide broad task and scene coverage, while ARX-data specifically increases coverage of physically executed non-monotonic behavior such as grasp failures, object drops, wrong-object interactions, inaccurate placements, re-grasping, and pose correction.
Unlike public datasets that mainly contain successful executions, ARX-data intentionally includes failed, regressive, and recovery trajectories so that \vlac\ can learn to distinguish harmful trial-and-error behavior from useful progress and recovery segments.
All selected videos are re-annotated using the same process-level schema; progress labels are not inherited from source datasets.
The curated training/evaluation dataset contains 28,167 video-level records, 22,978 episodes, 15,206 task units, and 375,172 progress points.
We reserve 3,515 records for held-out VPB evaluation and use the remaining training split to construct instruction-tuning data.
Appendix~\ref{app:vlac_dataset_details} provides the full source inventory, annotation protocol, split table, and dataset visualizations.

\subsection{Instruction Tuning}
\label{sec:training_data_and_model}

We convert the training split into multimodal conversations that preserve the process structure of the annotations.
Inputs may be a single image, an image pair, a sampled video clip, a rollout prefix, or optional robot/action context.
Targets include progress values, state and action descriptions, progress explanations, success/failure analysis, correction plans, grounding labels, and relative action deltas.

The annotation-derived instruction data are organized around four abilities.
\textbf{Task decomposition and grounding} connects language goals with semantic milestones, task-relevant objects, gripper positions, and keypoints.
\textbf{Temporal progress prediction} teaches the model to estimate timestamp-aligned progress from task-conditioned visual evidence rather than elapsed time.
\textbf{Diagnostic reasoning} turns scalar progress supervision into explanations of why a segment is progress-making, idle, failure-inducing, or recoverable.
\textbf{In-context and action-conditioned supervision} prepares \vlac\ for rollout segmentation by conditioning on reference episodes and, when robot trajectories are available, predicting relative action deltas over progress-point or kinematic-keyframe intervals.
\vlac\ remains a rollout segmentation critic rather than a closed-loop control policy.

We additionally use two targeted augmentations.
Counterfactual reverse-progress augmentation constructs regression-and-recovery examples from pairs of annotated progress points, discouraging the shortcut that later frames are always more complete.
Grounding-based rationale generation verbalizes geometric evidence from object boxes and gripper keypoints when it is consistent with the annotated progress direction.
The supervised fine-tuning corpus is centered on this annotation-derived robot data and complemented with public multimodal, robotics, progress-reasoning, and spatial/video reasoning datasets.
Implementation details and the full data mixture are provided in Appendix~\ref{app:vlac_training_details}.

\subsection{The Video Progress Benchmark}
\label{sec:vpb}

We introduce the Video Progress Benchmark (VPB) to evaluate whether a video-language model can recover task-conditioned progress on held-out robot trajectories.
VPB uses the held-out split described above and reserves all progress annotations strictly for evaluation.
The split is organized along two axes that match the intended use of \vlac: semantic task familiarity, separating seen and unseen task units, and progress-pattern type, separating expert-like trajectories from non-expert trajectories with regressive or recovery behavior.
This design tests whether a model can estimate progress under both familiar and held-out task semantics, and whether it remains robust to non-monotonic executions such as failed grasps, wrong-object interactions, object drops, and recovery attempts.

\label{sec:vpb_metrics}
VPB evaluates progress understanding at three complementary scopes.
\textbf{Global progress metrics} measure whether the model recovers the continuous progress trajectory, using Progress Rank Correlation (PRC), Value-Order Correlation (VOC) on expert trajectories, and Mean Absolute Error (MAE).
\textbf{Terminal-state metrics} evaluate whether the model recognizes whether the final state reaches near-completion, using terminal-state accuracy and macro F1.
\textbf{Local direction metrics} evaluate whether adjacent annotated keyframes are correctly classified as improvement, stagnation, or regression, which directly measures the effectiveness of video process segmentation (cutting).
Appendix~\ref{app:vpb_details} gives the formal task definition, interpolation rule, and metric equations.

%% file: sections/experiements_updated.tex
\section{Experiments}
\label{sec:experiments}

This section first evaluates \vlac\ as a process-level rollout segmentation critic on VPB, and then evaluates the full VLAC-Cut guided post-training pipeline in real-world robot tasks.

\subsection{VLAC-Cut Evaluation on VPB}
\label{sec:vlac_cut_experiments}

\subsubsection{Experimental Setup}
\label{sec:exp_setup}

We evaluate \vlac\ on the formal VPB evaluation split described in
Section~\ref{sec:vpb} with 3,515 held-out robot trajectory records.
All methods are evaluated under the same official VPB protocol, using the
three metric families defined in Section~\ref{sec:vpb_metrics}: global
progress prediction, which measures the accuracy of trajectory-level progress
estimation; terminal-state recognition, which evaluates whether the final
state is correctly identified as successful or failed/incomplete; and local
progress direction, which measures whether adjacent transitions reflect
progress or regression.

We compare \vlac\ against representative progress- and reward-estimation
baselines for robotic video understanding. ProgressLM-RL is included as a
single-observation progress-reasoning baseline: each VPB query point is paired
with a task demonstration and evaluated as an image-text progress estimation
problem, after which the pointwise predictions are assembled into trajectory
progress curves. Robometer and RoboReward represent video-based robotic reward
models. For VPB, they are evaluated on trajectory prefixes so that each prefix
yields a current progress estimate, which is then aligned with the official
annotation points. Robo-Dopamine is included as a process reward modeling
baseline based on relative progress estimation between earlier and later
states; its pairwise progress predictions are converted into the same signed
progress representation used by VPB. TOPReward is used as a zero-shot
token-probability reward baseline: we query each trajectory prefix and convert
its completion evidence into a normalized progress curve for evaluation.
We also evaluate Generative Value Learning (GVL) style baselines, which use a
VLM as an in-context progress estimator over selected keyframes. Since GVL is a
prompting framework rather than a single fixed model, we instantiate it with
three VLM backbones: GPT-5.5, Gemini-3.1-Pro, and Gemini-3.5-Flash. We report
two variants for each backbone. The first, denoted GVL, follows the original
shuffled-frame formulation, where keyframes are presented in a randomized order
and the model predicts their task-completion percentages. The second, denoted
Chrono-GVL, uses nearly identical keyframes and prompt structure but presents
frames in chronological order. In both cases, predictions on selected keyframes
are interpolated back to the full VPB timeline before computing metrics. This
design lets us compare the original temporal-ordering formulation of GVL with a
chronological variant while keeping the underlying VLM backbone explicit.

\subsubsection{Evaluation Results}
\label{sec:exp_results}

\paragraph{Global-Level Progress Prediction}
\label{sec:exp_curve}
Table~\ref{tab:curve_main} reports global progress prediction results on
the full VPB split and on the four evaluation buckets. \vlac\ achieves the
best overall MAE and PRC, with an MAE of 7.5600 and a PRC of 0.9260. It
also obtains the lowest MAE in every bucket. The strongest non-\vlac\
method under overall MAE is Chrono-GVL-GPT-5.5, with 12.3511, while the
strongest non-\vlac\ method under overall PRC is Chrono-GVL-Gemini-3.1-Pro,
with 0.8997. Thus, the gains are not confined to either value calibration
or temporal ordering: \vlac\ more accurately recovers both the absolute
signed progress values and the shape of the progress trajectory.

\begin{figure*}[t]
\centering
\includegraphics[width=1.0\linewidth]{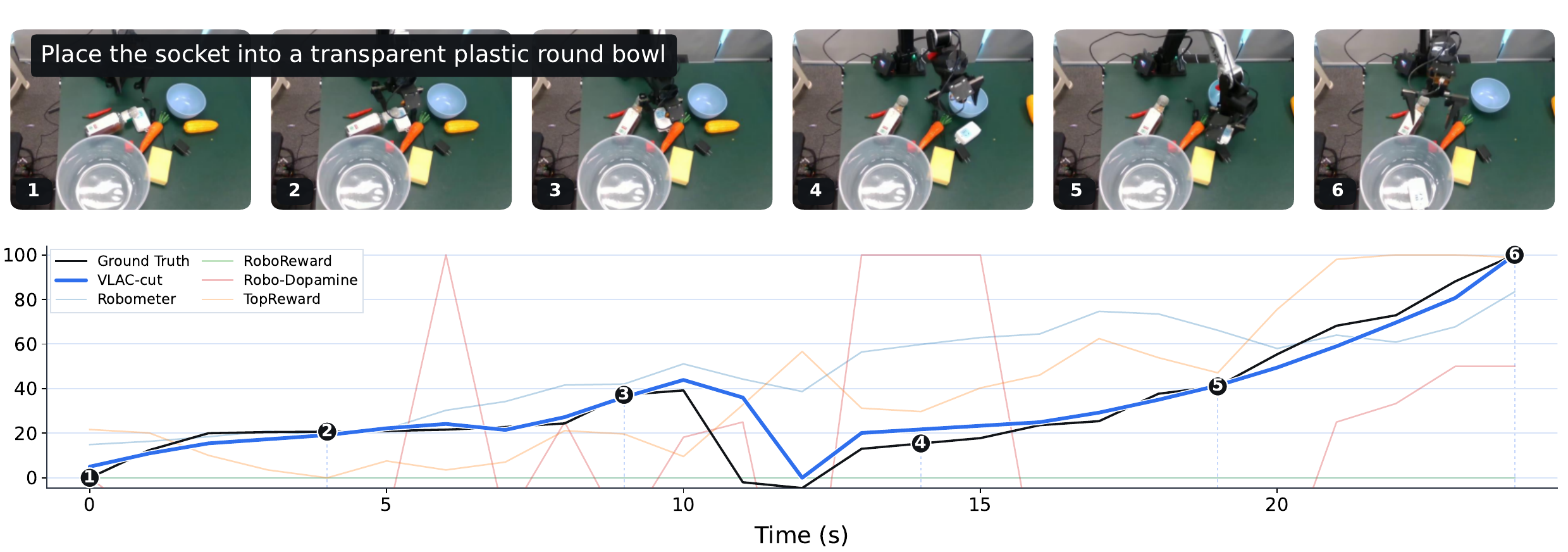}
\caption{
Qualitative comparison of signed progress prediction on a representative
VPB trajectory. The top row shows sampled video frames, and the numbered
markers link each frame to the corresponding timestamp on the progress
curves. The ground-truth trajectory contains an early regression followed
by gradual recovery and final task completion.
}
\label{fig:qualitative_progress_case}
\end{figure*}

\begin{table*}[tb]
\centering
\scriptsize
\setlength{\tabcolsep}{2.1pt}
\renewcommand{\arraystretch}{1.06}
\resizebox{\textwidth}{!}{%
\begin{tabular}{lcc ccc ccc cc cc}
\toprule
\multirow{2}{*}{Method}
& \multicolumn{2}{c}{Overall}
& \multicolumn{3}{c}{Expert Seen}
& \multicolumn{3}{c}{Expert Unseen}
& \multicolumn{2}{c}{Non-Expert Seen}
& \multicolumn{2}{c}{Non-Expert Unseen} \\
\cmidrule(lr){2-3}
\cmidrule(lr){4-6}
\cmidrule(lr){7-9}
\cmidrule(lr){10-11}
\cmidrule(lr){12-13}
& MAE $\downarrow$ & PRC $\uparrow$
& MAE $\downarrow$ & PRC $\uparrow$ & VOC $\uparrow$
& MAE $\downarrow$ & PRC $\uparrow$ & VOC $\uparrow$
& MAE $\downarrow$ & PRC $\uparrow$
& MAE $\downarrow$ & PRC $\uparrow$ \\
\midrule
\vlac
& \textbf{7.5600} & \textbf{0.9260}
& \textbf{6.3858} & \textbf{0.9910} & \textbf{0.9933}
& \textbf{6.0478} & 0.9859 & \textbf{0.9877}
& \textbf{9.5572} & \textbf{0.8284}
& \textbf{9.4845} & \textbf{0.8413} \\
ProgressLM-RL
& 30.7447 & 0.2083
& 30.5008 & 0.2174 & 0.2156
& 30.2872 & 0.2193 & 0.2175
& 31.0131 & 0.1963
& 31.4990 & 0.1907 \\
Robometer
& 21.1340 & 0.6605
& 20.0917 & 0.7542 & 0.7535
& 20.8336 & 0.7208 & 0.7195
& 22.2107 & 0.5416
& 22.0178 & 0.5548 \\
RoboReward
& 19.7949 & 0.7521
& 19.4609 & 0.8318 & 0.8302
& 21.3870 & 0.8155 & 0.8137
& 18.4670 & 0.6506
& 19.2844 & 0.6381 \\
Robo-Dopamine
& 29.5805 & 0.5604
& 27.6276 & 0.6539 & 0.6529
& 29.3321 & 0.6161 & 0.6162
& 30.4673 & 0.4572
& 31.9039 & 0.4456 \\
TOPReward
& 32.8329 & 0.2692
& 31.7592 & 0.2911 & 0.2854
& 31.1262 & 0.2982 & 0.2945
& 34.7699 & 0.2410
& 34.9541 & 0.2233 \\
GVL-GPT-5.5
& 25.1842 & 0.4029
& 24.3398 & 0.4362 & 0.4356
& 24.4317 & 0.4314 & 0.4309
& 26.6004 & 0.3414
& 26.1002 & 0.3737 \\
GVL-Gemini-3.1-Pro
& 25.3070 & 0.4600
& 22.5383 & 0.5215 & 0.5208
& 23.7037 & 0.4879 & 0.4863
& 28.6619 & 0.3879
& 28.3346 & 0.4010 \\
GVL-Gemini-3.5-Flash
& 22.8278 & 0.5370
& 19.7300 & 0.6091 & 0.6082
& 21.3906 & 0.5570 & 0.5558
& 25.9556 & 0.4767
& 26.3194 & 0.4626 \\
Chrono-GVL-GPT-5.5
& 12.3511 & 0.8768
& 10.0395 & 0.9771 & 0.9770
& 10.6762 & 0.9789 & 0.9793
& 15.3855 & 0.7214
& 15.1366 & 0.7354 \\
Chrono-GVL-Gemini-3.1-Pro
& 12.7982 & 0.8997
& 8.8988 & 0.9818 & 0.9818
& 9.4287 & \textbf{0.9869} & 0.9868
& 17.9306 & 0.7629
& 18.2758 & 0.7836 \\
Chrono-GVL-Gemini-3.5-Flash
& 12.8388 & 0.8961
& 9.0016 & 0.9868 & 0.9872
& 9.2796 & 0.9849 & 0.9852
& 18.1428 & 0.7636
& 18.3315 & 0.7643 \\
\bottomrule
\end{tabular}%
}
\caption{Global-level VPB results on the overall split and four evaluation
buckets. Lower MAE is better; higher PRC and VOC are better. VOC is reported
only for expert buckets, following the benchmark definition.}
\label{tab:curve_main}
\end{table*}

The expert buckets isolate progress estimation under trajectories that are
expected to be chronologically ordered. In this setting, \vlac\ obtains
MAE 6.3858, PRC 0.9910, and VOC 0.9933 on expert seen tasks, and MAE
6.0478, PRC 0.9859, and VOC 0.9877 on expert unseen tasks. Chrono-GVL
variants also achieve strong ordering performance, and
Chrono-GVL-Gemini-3.1-Pro slightly exceeds \vlac\ on expert-unseen PRC.
However, its MAE remains substantially higher, indicating that
chronological ordering alone is insufficient for calibrated signed-progress
prediction. The non-expert buckets are more challenging because trajectories
may regress or recover after failed intermediate actions. \vlac\ obtains
MAE 9.5572 and PRC 0.8284 on non-expert seen tasks, and MAE 9.4845 and PRC
0.8413 on non-expert unseen tasks, outperforming all baselines under both
metrics. Figure~\ref{fig:qualitative_progress_case} illustrates the same
behavior qualitatively: \vlac\ follows the annotated regression and recovery
pattern, whereas several baselines either produce unstable estimates or fail
to capture the nuances of the regressive dynamics.

\paragraph{Terminal-State Recognition}
\label{sec:exp_terminal}

Table~\ref{tab:terminal_main} reports terminal-state recognition on the
full VPB split and on the merged seen and unseen partitions. Terminal
recognition evaluates whether the final predicted progress is calibrated
to the task outcome. Since VPB contains both successful and
failed/incomplete terminal states, we report class-wise F1 together with
their macro average.

\begin{table*}[t]
\centering
\scriptsize
\setlength{\tabcolsep}{1.8pt}
\renewcommand{\arraystretch}{1.06}
\resizebox{\textwidth}{!}{%
\begin{tabular}{lcccc cccc cccc}
\toprule
\multirow{2}{*}{Method}
& \multicolumn{4}{c}{Overall}
& \multicolumn{4}{c}{Seen}
& \multicolumn{4}{c}{Unseen} \\
\cmidrule(lr){2-5}
\cmidrule(lr){6-9}
\cmidrule(lr){10-13}
& TSA $\uparrow$ & $\mathrm{F1}_{S}$ $\uparrow$ & $\mathrm{F1}_{F}$ $\uparrow$ & $\mathrm{MacroF1}_{T}$ $\uparrow$
& TSA $\uparrow$ & $\mathrm{F1}_{S}$ $\uparrow$ & $\mathrm{F1}_{F}$ $\uparrow$ & $\mathrm{MacroF1}_{T}$ $\uparrow$
& TSA $\uparrow$ & $\mathrm{F1}_{S}$ $\uparrow$ & $\mathrm{F1}_{F}$ $\uparrow$ & $\mathrm{MacroF1}_{T}$ $\uparrow$ \\
\midrule
\vlac
& 84.30 & 90.17 & \textbf{61.02} & \textbf{75.59}
& 83.77 & 89.76 & \textbf{60.91} & \textbf{75.33}
& \textbf{84.82} & 90.57 & \textbf{61.14} & \textbf{75.85} \\
ProgressLM-RL
& 18.63 & 4.41 & 29.17 & 16.79
& 18.34 & 4.65 & 28.59 & 16.62
& 18.93 & 4.17 & 29.75 & 16.96 \\
Robometer
& 24.78 & 18.04 & 30.49 & 24.27
& 24.60 & 18.17 & 30.10 & 24.13
& 24.96 & 17.91 & 30.89 & 24.40 \\
RoboReward
& 41.34 & 46.50 & 35.08 & 40.79
& 41.97 & 47.66 & 34.89 & 41.28
& 40.70 & 45.31 & 35.26 & 40.28 \\
Robo-Dopamine
& 47.23 & 56.67 & 32.52 & 44.59
& 46.87 & 56.38 & 32.05 & 44.21
& 47.58 & 56.96 & 32.99 & 44.98 \\
TOPReward
& 61.88 & 73.07 & 34.76 & 53.92
& 63.04 & 74.05 & 35.81 & 54.93
& 60.72 & 72.08 & 33.75 & 52.91 \\
GVL-GPT-5.5
& 51.92 & 62.36 & 33.46 & 47.91
& 51.37 & 62.01 & 32.44 & 47.22
& 52.47 & 62.71 & 34.48 & 48.60 \\
GVL-Gemini-3.1-Pro
& 51.24 & 62.45 & 30.49 & 46.47
& 52.33 & 63.78 & 30.31 & 47.04
& 50.14 & 61.07 & 30.67 & 45.87 \\
GVL-Gemini-3.5-Flash
& 55.90 & 67.63 & 30.87 & 49.25
& 57.00 & 68.84 & 30.67 & 49.76
& 54.80 & 66.38 & 31.05 & 48.72 \\
Chrono-GVL-GPT-5.5
& 67.82 & 77.86 & 41.12 & 59.49
& 69.31 & 79.23 & 41.22 & 60.23
& 66.34 & 76.45 & 41.04 & 58.74 \\
Chrono-GVL-Gemini-3.1-Pro
& 86.15 & 91.99 & 48.90 & 70.44
& 87.76 & 92.95 & 53.36 & 73.16
& 84.54 & 91.01 & 44.72 & 67.86 \\
Chrono-GVL-Gemini-3.5-Flash
& \textbf{86.69} & \textbf{92.48} & 42.08 & 67.28
& \textbf{88.72} & \textbf{93.65} & 49.49 & 71.57
& 84.65 & \textbf{91.30} & 35.10 & 63.20 \\
\bottomrule
\end{tabular}%
}
\caption{Terminal-state recognition results on VPB. Seen and unseen denote
the merged task-familiarity partitions. TSA measures terminal-state
accuracy; \(\mathrm{F1}_{S}\) and \(\mathrm{F1}_{F}\) are class-wise F1
scores for successful and failed/incomplete terminal states, respectively.
All values are reported in percent (\%).}
\label{tab:terminal_main}
\end{table*}

On the overall split, Chrono-GVL-Gemini-3.5-Flash obtains the highest
terminal-state accuracy and successful-terminal F1, with 86.69\% TSA and
92.48\% \(\mathrm{F1}_{S}\). In contrast, \vlac\ achieves the best
failed/incomplete-terminal F1 and the best terminal macro F1, with
\(\mathrm{F1}_{F}=61.02\%\) and \(\mathrm{MacroF1}_{T}=75.59\%\). This
indicates that chronological prompting is effective at recognizing visually
plausible successful endings, while \vlac\ provides a more balanced terminal
decision by substantially improving recognition of failed or incomplete final
states.

The same pattern largely holds across seen and unseen tasks. Chrono-GVL-Gemini-3.5-Flash
has the highest TSA and \(\mathrm{F1}_{S}\) on seen tasks and the highest
\(\mathrm{F1}_{S}\) on unseen tasks, whereas \vlac\ obtains the highest
\(\mathrm{F1}_{F}\) and \(\mathrm{MacroF1}_{T}\) in both partitions, as
well as the highest unseen TSA. The gap is most important for the
failed/incomplete class: \vlac\ reaches 60.91\% \(\mathrm{F1}_{F}\) on seen
tasks and 61.14\% on unseen tasks, outperforming all baselines. This
suggests that dense signed-progress supervision improves terminal-state
calibration beyond success-biased endpoint recognition.

\paragraph{Local Progress Direction}
\label{sec:exp_direction}

Table~\ref{tab:direction_main} reports local progress-direction AP on
adjacent semantic-anchor transitions. The metric uses the predicted local
progress difference as a ranking score for annotated increases, and its
negation as a ranking score for annotated decreases. We report
\(\mathrm{AP}_{+}\), \(\mathrm{AP}_{-}\), and their average
\(\mathrm{MacroAP}_{D}\).

\begin{table*}[t]
\centering
\scriptsize
\setlength{\tabcolsep}{2.8pt}
\renewcommand{\arraystretch}{1.06}
\resizebox{\textwidth}{!}{%
\begin{tabular}{lccc ccc ccc}
\toprule
\multirow{2}{*}{Method}
& \multicolumn{3}{c}{Overall}
& \multicolumn{3}{c}{Seen}
& \multicolumn{3}{c}{Unseen} \\
\cmidrule(lr){2-4}
\cmidrule(lr){5-7}
\cmidrule(lr){8-10}
& $\mathrm{AP}_{+}$ $\uparrow$
& $\mathrm{AP}_{-}$ $\uparrow$
& $\mathrm{MacroAP}_{D}$ $\uparrow$
& $\mathrm{AP}_{+}$ $\uparrow$
& $\mathrm{AP}_{-}$ $\uparrow$
& $\mathrm{MacroAP}_{D}$ $\uparrow$
& $\mathrm{AP}_{+}$ $\uparrow$
& $\mathrm{AP}_{-}$ $\uparrow$
& $\mathrm{MacroAP}_{D}$ $\uparrow$ \\
\midrule
\vlac
& \textbf{95.90} & \textbf{39.46} & \textbf{67.68}
& \textbf{96.30} & \textbf{38.34} & \textbf{67.32}
& \textbf{95.53} & \textbf{40.61} & \textbf{68.07} \\
ProgressLM-RL
& 91.41 & 9.35 & 50.38
& 92.69 & 9.52 & 51.10
& 90.22 & 9.27 & 49.74 \\
Robometer
& 93.41 & 11.46 & 52.44
& 94.19 & 11.21 & 52.70
& 92.66 & 11.73 & 52.19 \\
RoboReward
& 92.52 & 15.02 & 53.77
& 93.67 & 14.67 & 54.17
& 91.35 & 15.45 & 53.40 \\
Robo-Dopamine
& 92.07 & 10.12 & 51.09
& 92.94 & 9.94 & 51.44
& 91.20 & 10.34 & 50.77 \\
TOPReward
& 93.02 & 10.24 & 51.63
& 93.69 & 9.89 & 51.79
& 92.36 & 10.64 & 51.50 \\
GVL-GPT-5.5
& 92.24 & 9.11 & 50.68
& 93.41 & 9.45 & 51.43
& 91.11 & 8.87 & 49.99 \\
GVL-Gemini-3.1-Pro
& 92.19 & 9.83 & 51.01
& 92.78 & 9.52 & 51.15
& 91.62 & 10.16 & 50.89 \\
GVL-Gemini-3.5-Flash
& 92.67 & 10.77 & 51.72
& 93.65 & 11.27 & 52.46
& 91.70 & 10.44 & 51.07 \\
Chrono-GVL-GPT-5.5
& 94.83 & 16.00 & 55.42
& 95.23 & 16.59 & 55.91
& 94.41 & 15.49 & 54.95 \\
Chrono-GVL-Gemini-3.1-Pro
& 93.65 & 17.92 & 55.78
& 94.02 & 17.23 & 55.62
& 93.27 & 18.52 & 55.90 \\
Chrono-GVL-Gemini-3.5-Flash
& 94.40 & 22.00 & 58.20
& 94.63 & 21.32 & 57.98
& 94.16 & 22.71 & 58.43 \\
\bottomrule
\end{tabular}%
}
\caption{Local progress-direction AP on adjacent semantic-anchor
transitions. \(\mathrm{AP}_{+}\) ranks annotated progress increases using
the predicted local progress difference, while \(\mathrm{AP}_{-}\) ranks
annotated progress decreases using the negated predicted local progress
difference. All values are reported in percent (\%).}
\label{tab:direction_main}
\end{table*}

\vlac\ achieves the strongest local direction ranking performance across
the overall, seen, and unseen partitions. On the overall split, it obtains
\(\mathrm{AP}_{+}=95.90\%\), \(\mathrm{AP}_{-}=39.46\%\), and
\(\mathrm{MacroAP}_{D}=67.68\%\). The strongest non-\vlac\ macro score is
58.20\% from Chrono-GVL-Gemini-3.5-Flash. The largest gap appears in
\(\mathrm{AP}_{-}\), where \vlac\ substantially exceeds the strongest
non-\vlac\ score of 22.00\%. This indicates that \vlac\ provides a much
stronger ranking signal for regressive local changes, rather than merely
following chronological progress.

The seen and unseen results show the same trend. \vlac\ obtains
\(\mathrm{MacroAP}_{D}=67.32\%\) on seen tasks and 68.07\% on unseen tasks,
remaining ahead of all baselines in both partitions. Although chronological
variants achieve high \(\mathrm{AP}_{+}\), their \(\mathrm{AP}_{-}\) remains
substantially lower, indicating that chronological ordering alone is
insufficient for ranking regressive semantic-anchor transitions. This
highlights the value of process-level signed progress supervision for
capturing local task-state reversals.

%% file: sections/experiments_zhr.tex
\subsection{Real-World Post-Training Experiments}
\label{sec:real_world_experiments}
Our real-world post-training experiments are conducted under the two-operator, twelve-robot HELP deployment. This role-specialized supervision setup already provides high system-level human efficiency by allowing limited human labor to oversee concurrent robot interactions, expose recurring policy failure modes, and collect targeted recovery demonstrations. On top of this shared operating setup, we evaluate whether \vlac\ further improves the policy gain obtained from the same HITL recovery supervision. We adopt Human-in-the-Loop (HITL) training as the baseline: when a rollout enters an error-prone state, the operator provides a corrective demonstration to recover the robot, producing targeted recovery data for policy improvement. We therefore compare HELP with HITL-only updates when both start from the same checkpoint and use the same number of HITL recovery trajectories, and quantify this effect using a human-supervision amplification factor.

All real-world experiments use $\pi_{0.5}$ VLA~\cite{intelligence2025pi_}, and standard flow-matching objective is adopted to optimize the policy.

\subsubsection{Evaluation Tasks}
\begin{figure}[t]
    \centering
    \captionsetup{font=small,skip=3pt}
    \includegraphics[width=.88\linewidth]{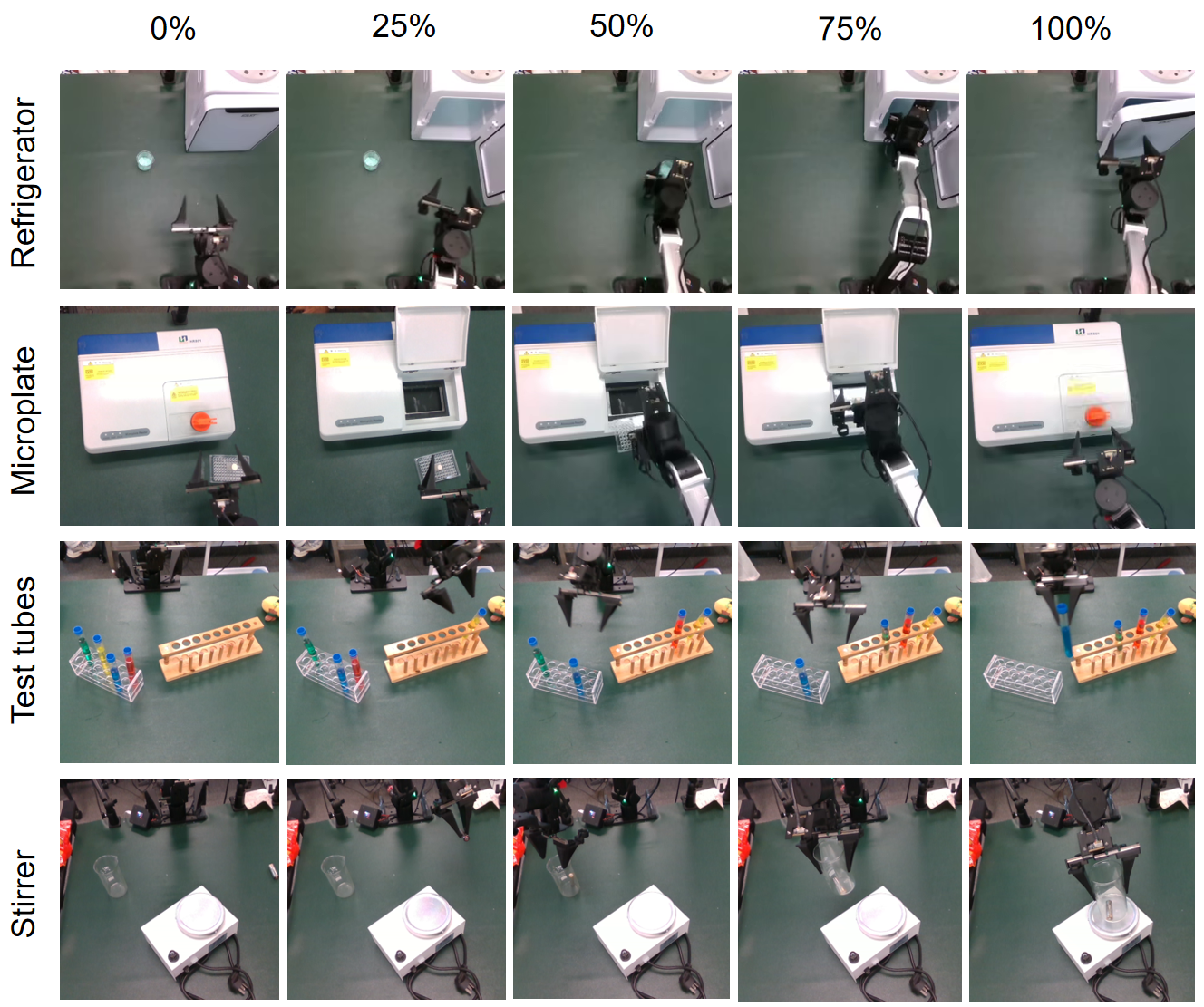}
    \caption{\textbf{Task execution progress.} Each row shows execution states for tasks, 
    and each column corresponds to a normalized task-completion progress from 0\% to 100\%.}
    \label{fig:task_progress}
\end{figure}

To evaluate policy performance, 
we conducted a comprehensive evaluation across four diverse real-world tasks: 
\textbf{Refrigerator}, \textbf{Microplate}, \textbf{Test Tube}, 
and \textbf{Stirrer}.
We summarize the tasks below, with progress illustrations in Figure~\ref{fig:task_progress}:

\textbf{Refrigerator}: 
In this task, the robot is required to open the refrigerator door, grasp a reagent-filled beaker with the gripper, place it stably inside the refrigerator, and then close the door.
For quantitative evaluation, we count a trial as successful only when the robot completes
the full task within 100 seconds without dropping the beaker or spilling any liquid during transfer. Although the policy may still recover by picking up the beaker after a drop, such cases are counted as failures to ensure consistent evaluation and reflect practical use. We use the following prompt as the instruction input to the VLA models for this task: \textit{Open the refrigerator door, grasp the beaker with the gripper, place it inside the refrigerator, and then close the door.}

\textbf{Microplate}: 
This task requires the robot to open the lid of the microplate reader, pick up a microplate from an arbitrary position on the table, transfer it into the reader, and close the lid. To be counted as successful, the robot must finish the full sequence within 100 seconds without dropping the microplate during transfer. Prompt: \textit{Turn on the microplate reader, place the microplate inside, then close the microplate reader.}

\textbf{Test Tube}: 
We evaluate our policies on the challenging long-horizon task of transferring four reagent-filled test tubes from a transparent plastic rack to a yellow wooden rack. Since the policy may succeed after multiple attempts during test-tube insertion, we set the time limit to 200 seconds for quantitative evaluation. Dropping any test tube during the transfer is also counted as a failure. Prompt: \textit{Move the test tube from the transparent rack to the yellow wooden rack}

\textbf{Stirrer}: 
This task involves grasping a magnetic stir bar from an arbitrary position on the table,
placing it into a beaker containing liquid, and then transferring the beaker onto a magnetic stirrer. A trial is considered successful only if the full task is completed within 60 seconds and no liquid is spilled during the transfer. Prompt: \textit{Grasp the magnetic stir bar from the desktop, place it into the beaker, and then put the beaker onto the magnetic stirrer.}

\subsubsection{Quantitative Results}

We use three metrics to evaluate the performance of models across different tasks: throughput, success rate, and failure progress. Throughput is defined as the number of tasks successfully completed per hour, measuring both execution efficiency and task completion capability. Success rate measures the proportion of episodes in which the policy successfully completes the task. We further evaluate failure progress, defined as the proportion of the task successfully completed before failure. This metric reflects a policy's potential to complete the task as well as its robustness, providing additional performance information beyond the binary success/failure outcome of each episode.

\paragraph{Iterative policy improvement with HELP.}

We first describe how HELP improves the policy through multiple rounds of data collection and training. In each round, HELP collects both HITL recovery data and autonomous rollout trajectories. \vlac\ segments the autonomous rollouts and extracts progress-making and recovery segments for the next policy update. As described in Section~\ref{sec:framework}, each iteration uses the best-performing model from the previous iteration to collect these data. We then construct a new training dataset by combining the curated rollout data, the HITL recovery data, and the data used to train the base model, approximately balancing them at a 1:1:1 ratio to achieve strong model performance. Finally, we train the current inference model on this combined dataset to obtain an updated policy, and the best-performing checkpoint is used as the inference model for the next round of data collection and training.

For the Refrigerator, Microplate, and Test Tube tasks, we conducted two iterations of data collection and training to improve the policy. In contrast, the Stirrer task was relatively simple, with the policy reaching a 90\% success rate after a single iteration;
therefore, no additional iteration was performed for this task.


Table \ref{tab:results} summarizes the policy evaluation results. After multiple iterations of training, HELP achieves substantial improvements in both throughput and success rate across the four evaluation tasks. Compared with the base model, the final policy improves these metrics by more than two times in most cases and by nearly three times in some tasks. In the final iteration, HELP achieves a success rate above 90\% in all tasks, except for refrigerator opening. The success rate of the refrigerator-opening task saturates at 80\%, likely because further improvement is constrained by the camera viewpoint and the mechanical structure. These results demonstrate that HELP can effectively improve policy performance.

A closer analysis of the same results indicates that the throughput improvement is primarily driven by the increase in success rate, while the execution time first increases and then decreases over the two optimization iterations. For the three tasks with a second round of optimization, the first iteration substantially improves the success rate over the base model, but it also introduces a noticeable increase in execution time. This is mainly because the added HITL recovery data teaches the policy to recover from states that previously led to failures. When the policy enters these error-prone states, it often performs additional recovery actions before returning to the correct execution trajectory. As a result, the initial gain in success rate is achieved largely by correcting erroneous steps, at the cost of longer execution time.

However, in the final iteration, throughput improves more substantially than in the first iteration. For the refrigerator task, throughput increases from 10 to 19 after the first iteration and further rises to 42 after the second iteration, corresponding to gains of 90\% and 120\%. Similar trends are observed in the Microplate task, where throughput increases from 13 to 18 and then to 42, with gains of 38\% and 133\%, respectively. In the Test Tube task, throughput increases from 14 to 20 and then to 37, corresponding to gains of 43\% and 83\%. This larger second-iteration improvement occurs because the stronger policy produces more successful autonomous rollout trajectories, which provide higher-quality data for the next round of training. Training on the curated rollout data enables the policy to execute correct actions more directly, rather than frequently entering error-prone states and relying on recovery behaviors as in the first iteration. Consequently, both execution time and success rate improve, leading to a larger throughput gain in the second iteration.

The quantitative definition of task progress for each task is illustrated in Figure~\ref{fig:task_progress}. The trend in failure progress provides further evidence of improved policy robustness, as the policy is able to complete a larger portion of each task even when it eventually fails. In the final Microplate iteration, the remaining failures are mainly concentrated at the most difficult step, opening the microplate reader lid. Once this step is completed, the subsequent actions are typically successful. This bottleneck explains the relatively low average failure progress of 20\%, since failed trials tend to terminate at the same early stage.

Overall, these results show that HELP substantially improves both success rate and throughput across most tasks, demonstrating its effectiveness in improving policy performance through iterative data collection and training.

\paragraph{Human-supervision amplification under matched HITL budgets.}


For an evaluation metric $M$, we define the human-supervision amplification factor as
\begin{equation}
A_M =
\frac{M(\pi_{\mathrm{HELP}})-M(\pi_{\mathrm{start}})}
{M(\pi_{\mathrm{HITL}})-M(\pi_{\mathrm{start}})}.
\end{equation}
Here, $\pi_{\mathrm{start}}$ is the common starting checkpoint for the matched comparison, while $\pi_{\mathrm{HITL}}$ and $\pi_{\mathrm{HELP}}$ are the policies after HITL-only and HELP updates, respectively. Because both methods use the same number of HITL recovery trajectories within each task-round, $A_M$ is equivalent to the ratio of performance improvement per HITL recovery trajectory. A value $A_M>1$ indicates that \vlac{}-curated autonomous rollouts produce a larger policy gain from the same HITL recovery budget.

We compare HELP with multi-iteration HITL training under matched HITL recovery budgets. For each task and training iteration, both methods start from the same policy checkpoint and use the same number of HITL recovery trajectories, as summarized in Table~\ref{tab:episodes}. HELP additionally reuses autonomous rollout segments curated by \vlac. Table~\ref{tab:amplification} reports the resulting human-supervision amplification factors derived from the absolute policy results in Table~\ref{tab:results}. Across all seven matched task-iteration comparisons, the throughput-gain amplification factor ranges from 1.20$\times$ to 3.43$\times$, while the success-rate-gain amplification factor ranges from 1.50$\times$ to 3.00$\times$. The arithmetic means across the seven comparisons are 2.15$\times$ for throughput gain and 2.21$\times$ for success-rate gain. The amplification is observed across all tasks and in both the first and second post-training iterations.

These results quantify the learning-level component of human efficiency: HELP consistently outperforms HITL-only training under matched HITL recovery budgets. This improvement is enabled by \vlac, which extracts useful progress-making and recovery segments from autonomous rollouts without increasing the amount of human recovery supervision. These comparisons isolate the additional learning-level benefit of \vlac, since both settings already use the same two-operator, twelve-robot HELP supervision configuration. Multi-robot supervision first provides system-level human efficiency and failure-mode discovery, while \vlac\ further improves the utilization of the resulting rollout data. Together, these two levels of improvement support HELP as an overall human-efficient post-training pipeline.

\begin{table}[!ht]
    \centering
    \footnotesize
    \captionsetup{font=small,skip=3pt}
    \setlength{\tabcolsep}{4pt}
    \renewcommand{\arraystretch}{.86}
    \begin{tabular}{llcccc}
    \toprule[1.2pt]
    \textbf{Task} & \textbf{Model} & \textbf{Throughput} & \textbf{Success rate} & \textbf{Execution Time (s)} & \textbf{Failure progress} \\
    \midrule
    \multirow{5}{*}{Refrigerator}
        & Base model & 10 & 20\% & 72.3 & 40\% \\
        & HITL(i=1) & 14 & 35\% & 89 & 42.3\% \\
        & HELP(i=1) & 19 & 45\% & 85 & 49\% \\
        & HITL(i=2) & 33 & 60\% & 65.6 & 56\% \\
        & HELP(i=2) & 42 & 80\% & 67.2 & 62.5\% \\
    \midrule
    \multirow{5}{*}{Microplate}
        & Base model & 13 & 30\% & 83.3 & 30\% \\
        & HITL(i=1) & 16 & 40\% & 88.35 & 27.5\% \\
        & HELP(i=1) & 18 & 45\% & 91.4 & 34.5\% \\
        & HITL(i=2) & 25 & 60\% & 85.1 & 51.3\% \\
        & HELP(i=2) & 42 & 90\% & 77.7 & 20\% \\
    \midrule
    \multirow{5}{*}{Test Tube}
        & Base model & 14 & 35\% & 93 & 38.9\% \\
        & HITL(i=1) & 19 & 45\% & 83 & 41.2\% \\
        & HELP(i=1) & 20 & 55\% & 105 & 45\% \\
        & HITL(i=2) & 28 & 70\% & 90.7 & 65\% \\
        & HELP(i=2) & 37 & 95\% & 93.4 & 76\% \\
    \midrule
    \multirow{3}{*}{Stirrer}
        & Base model & 66 & 55\% & 30 & 33.3\% \\
        & HITL(i=1) & 83 & 70\% & 30.5 & 35\% \\
        & HELP(i=1) & 112 & 90\% & 29 & 45\% \\
    \bottomrule[1.2pt]
    \end{tabular}
    \caption{\textbf{Real-world post-training performance of HELP across iterative training rounds.}
    We report the throughput, success rate, average execution time, and failure progress for the 
    base model, HITL-only updates, and HELP.
    For each matched task-round comparison, HITL and HELP start from the same checkpoint.}
    \label{tab:results}
\end{table}

\begin{table}[!ht]
    \centering
    \footnotesize
    \captionsetup{font=small,skip=3pt}
    \setlength{\tabcolsep}{4.5pt}
    \renewcommand{\arraystretch}{.86}
    \begin{tabular}{llccc}
    \toprule[1.2pt]
    \textbf{Task} & \textbf{Model} & \textbf{Base demos.} & \textbf{HITL recovery} & \textbf{Curated rollout} \\
    \midrule
    \multirow{5}{*}{Refrigerator}
        & Base model & 94 & - & - \\
        & HITL(i=1) & 94 & 201 & - \\
        & HELP(i=1) & 94 & 201 & 100 \\
        & HITL(i=2) & 94 & 242 & - \\
        & HELP(i=2) & 94 & 242 & 117 \\
    \midrule
    \multirow{5}{*}{Microplate}
        & Base model & 112 & - & - \\
        & HITL(i=1) & 112 & 168 & - \\
        & HELP(i=1) & 112 & 168 & 173 \\
        & HITL(i=2) & 112 & 134 & - \\
        & HELP(i=2) & 112 & 134 & 116 \\
    \midrule
    \multirow{5}{*}{Test Tube}
        & Base model & 150 & - & - \\
        & HITL(i=1) & 150 & 352 & - \\
        & HELP(i=1) & 150 & 352 & 142 \\
        & HITL(i=2) & 150 & 300 & - \\
        & HELP(i=2) & 150 & 300 & 141 \\
    \midrule
    \multirow{3}{*}{Stirrer}
        & Base model & 101 & - & - \\
        & HITL(i=1) & 101 & 230 & - \\
        & HELP(i=1) & 101 & 230 & 334 \\
    \bottomrule[1.2pt]
    \end{tabular}
    \caption{\textbf{Training-data composition for each task and iteration.} 
    Each entry denotes the number of base demonstrations, HITL recovery trajectories, or curated rollout segments used for training. 
    Within each matched task-round comparison, HITL and HELP use the same number of base demonstrations and HITL recovery trajectories; HELP additionally uses \vlac{}-curated autonomous rollout segments.}
    \label{tab:episodes}
\end{table}

\begin{table}[!ht]
    \centering
    \scriptsize
    \captionsetup{font=small,skip=3pt}
    \setlength{\tabcolsep}{2.2pt}
    \renewcommand{\arraystretch}{.9}
    \resizebox{\linewidth}{!}{%
    \begin{tabular}{llccccccc}
    \toprule[1.2pt]
    \textbf{Task} & \textbf{Round} & \textbf{HITL budget} & $\Delta$\textbf{TP HITL} & $\Delta$\textbf{TP HELP} & $A_{\mathrm{TP}}$ & $\Delta$\textbf{SR HITL} & $\Delta$\textbf{SR HELP} & $A_{\mathrm{SR}}$ \\
    \midrule
    Refrigerator & 1 & 201 & +4 & +9 & 2.25$\times$ & +15 pp & +25 pp & 1.67$\times$ \\
    Microplate & 1 & 168 & +3 & +5 & 1.67$\times$ & +10 pp & +15 pp & 1.50$\times$ \\
    Test Tube & 1 & 352 & +5 & +6 & 1.20$\times$ & +10 pp & +20 pp & 2.00$\times$ \\
    Stirrer & 1 & 230 & +17 & +46 & 2.71$\times$ & +15 pp & +35 pp & 2.33$\times$ \\
    Refrigerator & 2 & 242 & +14 & +23 & 1.64$\times$ & +15 pp & +35 pp & 2.33$\times$ \\
    Microplate & 2 & 134 & +7 & +24 & 3.43$\times$ & +15 pp & +45 pp & 3.00$\times$ \\
    Test Tube & 2 & 300 & +8 & +17 & 2.13$\times$ & +15 pp & +40 pp & 2.67$\times$ \\
    \midrule
    \textbf{Arithmetic mean} & -- & -- & -- & -- & \textbf{2.15$\times$} & -- & -- & \textbf{2.21$\times$} \\
    \bottomrule[1.2pt]
    \end{tabular}%
    }
    \caption{\textbf{Comparison between HELP and HITL-only training under matched HITL recovery budgets.}
    For each task-round, HITL and HELP start from the same policy checkpoint and use the same number of HITL recovery trajectories. $\Delta$ denotes the performance improvement relative to the common starting checkpoint. $A_{\mathrm{TP}}$ and $A_{\mathrm{SR}}$ denote the amplification factors for throughput gain and success-rate gain, respectively. The final row reports the arithmetic mean across the seven matched task-round comparisons. This table quantifies supervision utilization and complements the system-level two-operator, twelve-robot result.}
    \label{tab:amplification}
\end{table}
\FloatBarrier

%% file: sections/contribution.tex
\section{Author Contribution Statement}

The authors confirm their contributions as follows:

\begin{itemize}[labelindent=0pt, leftmargin=*]
    \item Shaopeng Zhai (zsp1197@163.com): led the project; designed the algorithm and technical roadmap; developed the infrastructure; and contributed to data collection, real-world experiments, \vlac\, and policy post-training.
    \item Qi Zhang (zhangqi.fqz@gmail.com): led the development of \vlac\, including data collection, annotation, process, model training and VPB benchmark.
    \item Tianyi Zhang (tianyizhang0729@163.com): contributed to real-world experiments, infrastructure development, \vlac\ data annotation, policy post-training, and data collection.
    \item Haoran Zhang (zhrecnucee@gmail.com): VLA deployment and data collection, contributed substantially to policy post-training and eaperiments.
    \item Fuxian Huang (hfuxian@zju.edu.cn): data collection and contributed substantially to \vlac\ data annotation.
    \item Zhanhui Lin (zhanhuilin@link.cuhk.edu.cn): contributed to \vlac\ data preprocessing, collection, annotation, and model training; and contributed substantially to the VPB benchmark.
    \item Zijun Xu (25113070118@m.fudan.edu.cn): contributed to teleoperation system development.
\end{itemize}

%% file: sections/appendix_vlac.tex
\section{Additional Details for \vlac{}}
\label{app:vlac_cut}

\subsection{Progress Annotation Dataset Details}
\label{app:vlac_dataset_details}

The full annotation inventory is constructed from public robot datasets and a targeted in-house real-world collection.
DROID~\citep{khazatsky2024droid} contributes diverse real-world manipulation videos, while LIBERO~\citep{liu2023libero} and VLABench~\citep{zhang2025vlabench} provide simulated task variation and benchmark-style manipulation scenarios.
From these datasets, we extract task instructions, camera videos, trajectory metadata, and semantic task information, and then re-annotate selected videos with our progress labels.

Although these public datasets expand task and scene coverage, they are largely demonstration-centric: most trajectories correspond to successful or near-expert executions.
This distribution is insufficient for learning a general-purpose progress critic that must evaluate policy behavior beyond nominal demonstrations.
A critic trained primarily on expert-like trajectories may implicitly assume that execution quality improves monotonically with time and fail to distinguish partial progress, stagnation, regression, recoverable errors, and unrecoverable failure.
Reverse or rewound video augmentation can expose models to synthetic decreasing-progress examples~\citep{zhang2025rewind}, but it cannot fully substitute for physically executed non-expert robot rollouts with realistic contact dynamics and out-of-expert-distribution failure modes.

\begin{figure}[t]
    \centering
    \includegraphics[width=0.95\linewidth]{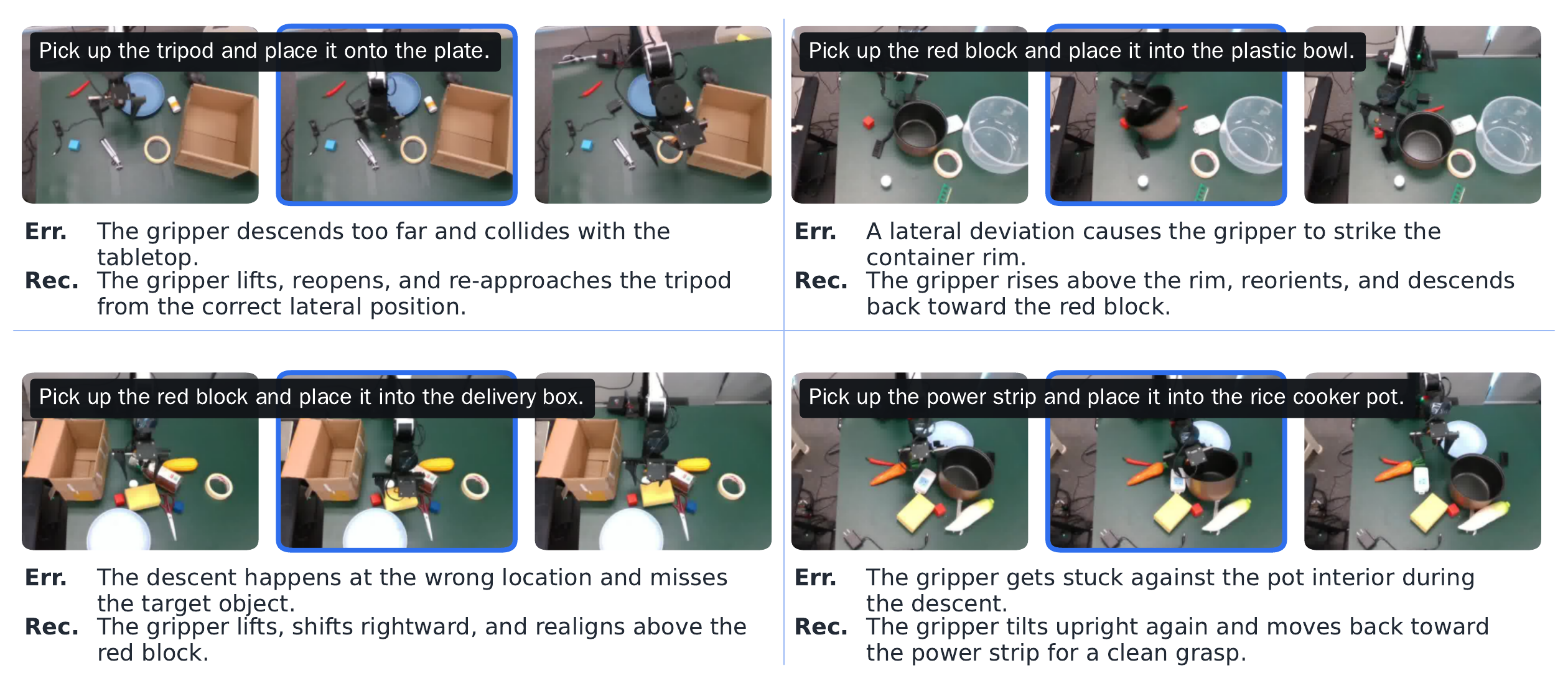}
    \caption{Representative failure (Err.) and recovery modes (Rec.) in ARX-data. These examples provide visual evidence for annotating stagnation, regression, and recovery in signed progress trajectories.}
    \label{fig:arx_failure_modes}
\end{figure}

To address this limitation, we collect ARX-data, a targeted in-house real-world robot dataset designed to increase coverage of physically executed non-monotonic progress.
ARX-data is predominantly collected on a single-robot real-world platform built around an ARX-5 robot under master--slave teleoperation, with synchronized multi-view videos.
In addition to successful demonstrations, the collection includes deliberately induced deviations from nominal execution, covering grasp failures, workspace collisions, unstable contacts, wrong-object interactions, inaccurate placements, object drops, deviations from the intended motion, and recovery behaviors such as re-grasping or correcting an object pose after failure, as displayed in Figure~\ref{fig:arx_failure_modes}.
These trajectories provide physically grounded examples in which task progress may increase, stagnate, decrease, or later recover.

Overall, the source design prioritizes real-world robot behavior.
DROID provides large-scale real-world manipulation diversity, while ARX-data complements it with physically executed non-monotonic progress trajectories.
LIBERO and VLABench are retained as supplementary simulated sources for controlled task variation and additional manipulation coverage, rather than as the main basis of the dataset.
Progress labels are not inherited from any source dataset; all selected videos are re-annotated using our process-level schema to produce view-conditioned signed progress supervision.

\begin{figure}[t]
    \centering
    \includegraphics[width=0.95\linewidth]{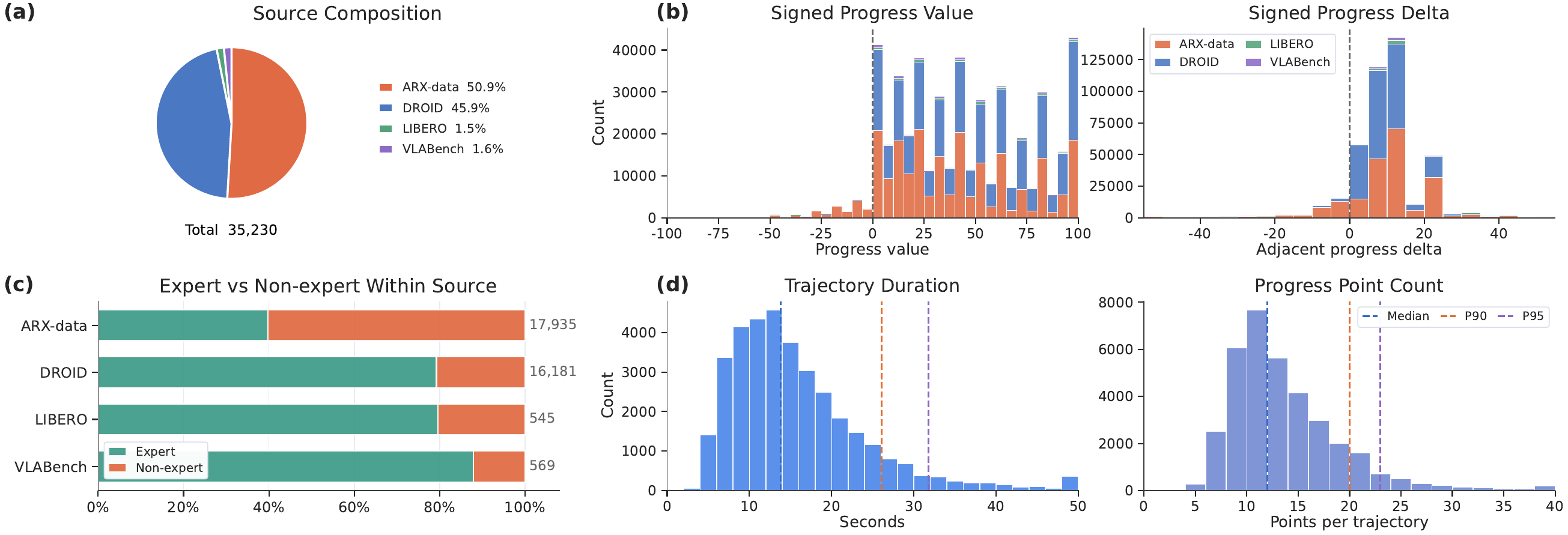}
    \caption{Full annotation inventory statistics. (a) Source composition and execution-quality distribution of the annotated robot video corpus. ARX-data and DROID dominate the real-world portion, while LIBERO and VLABench provide supplementary simulated coverage. (b) Signed progress statistics across data sources. The distribution of progress values and adjacent progress deltas shows that the corpus contains not only forward progress but also stagnation, regression, and recovery behavior. (c) Temporal characteristics of the progress annotations. Most trajectories are short manipulation episodes, while each episode contains multiple progress points that provide dense process-level supervision.}
    \label{fig:dataset_composition}
\end{figure}

\begin{figure}[t]
    \centering
    \includegraphics[width=0.86\linewidth]{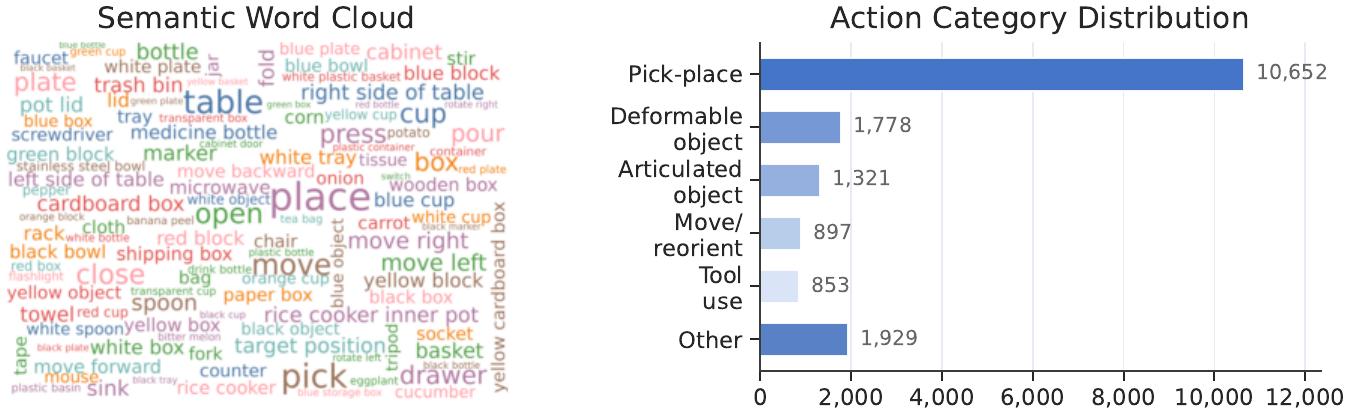}
    \caption{Semantic and action diversity of the full annotation inventory. The corpus covers a broad range of manipulation objects, spatial relations, and action categories, with pick-and-place tasks forming the largest group.}
    \label{fig:dataset_semantic_space}
\end{figure}

The full annotation inventory contains 35,230 records, 26,615 episodes, 15,206 task units, and 464,446 progress points.
A record denotes one annotated video-level sample and is the basic unit of progress annotation.
An episode denotes one physical task execution and may correspond to multiple camera- or view-specific records.
A task unit denotes a semantic manipulation task used for measuring task coverage and organizing benchmark splits.
For progress-pattern analysis, a non-expert record is defined as one whose ordered progress points contain at least one adjacent decrease in signed progress; otherwise it is expert.
Under this annotation-derived criterion, the full inventory contains 20,885 expert records and 14,345 non-expert records.
Figures~\ref{fig:dataset_composition} and~\ref{fig:dataset_semantic_space} summarize source composition, execution-quality distribution, signed progress statistics, temporal annotation density, and semantic/action diversity.

\subsection{Annotation Schema and Protocol}
\label{app:annotation_schema_protocol}

Each progress point is task-conditioned rather than time-conditioned: the same visual motion may indicate progress for one instruction and irrelevant or harmful behavior for another.
Annotators therefore mark sparse keyframes where task state changes, including forward progress, stagnation, regression, and recovery.
For each episode, the annotator first reviews the task instruction and creates a task-level plan.
The plan decomposes the instruction into semantic milestones that describe the nominal progression of the task, typically spanning the 0--100\% completion range.
This plan provides a task-conditioned reference for assigning signed progress values, but pointwise progress labels themselves are not restricted to monotonic completion.
Progress points may take negative values when the execution moves away from the goal or interacts with an incorrect object.

The annotator then marks progress points at meaningful timestamps.
A progress point is not a uniformly sampled frame; it is a semantic event where the task state changes, a subgoal is achieved, an error occurs, the execution stagnates, or the trajectory reaches an inflection point.
Annotators are instructed to identify moments where progress reverses, such as failed grasps, object drops, incorrect placements, or recovery attempts.
These inflection points define the temporal boundaries at which a trajectory can transition from improvement to stagnation, regression, or recovery.

\begin{figure}[t]
    \centering
    \includegraphics[width=0.95\linewidth]{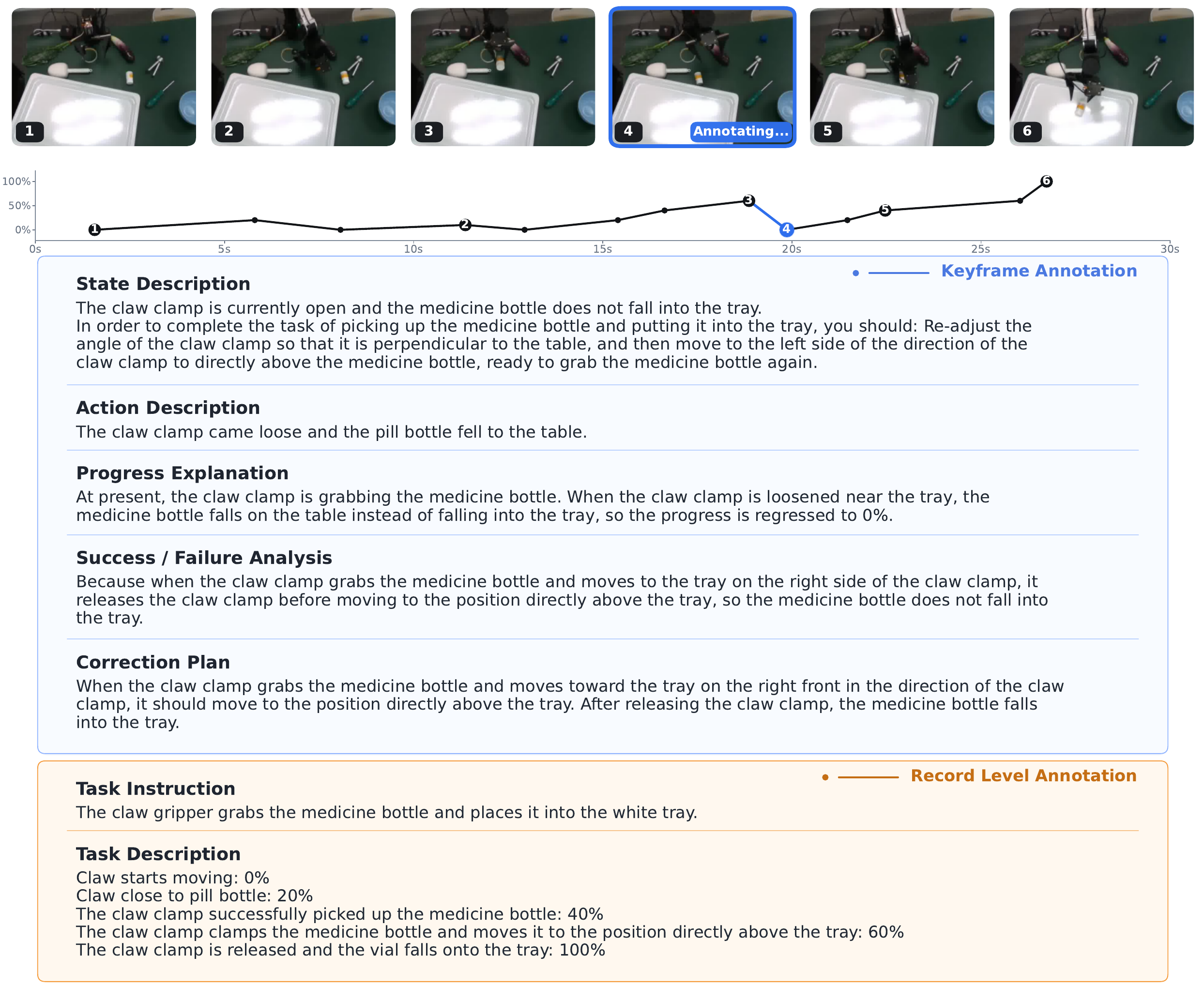}
    \caption{Example of sparse progress annotation for a robot manipulation trajectory. Only selected keyframes are manually annotated with signed progress values and structured diagnostic supervision, including state description, action description, progress explanation, success/failure analysis, and correction planning. Intermediate video frames are shown as unannotated temporal context, while the current keyframe corresponds to the active annotation target.}
    \label{fig:dataset_annotation}
\end{figure}

Each progress point aligns visual time with structured language supervision and optional spatial grounding.
The textual annotation contains five complementary components: state description, action description, progress explanation, success/failure analysis, and correction plan.
Optional grounding annotations localize the gripper end-effector and task-relevant objects on the keyframe corresponding to each progress point.
This grounding links language explanations to visual evidence and supports models that reason jointly about object state, robot motion, and task progress.
Figure~\ref{fig:dataset_annotation} summarizes the annotation schema.

We use an interactive annotation interface to support process-level labeling at scale.
The interface allows annotators to inspect multi-view videos, edit the task plan, mark progress points, assign signed progress values, and provide corresponding state, action, explanation, success/failure, correction, and grounding annotations.
For datasets that provide robot trajectories or action information, the system can initialize candidate progress points using kinematic analysis.
The kinematic mode loads robot positions and gripper states, rescales position, orientation, and gripper dimensions, unwraps rotations, filters trajectories, computes velocity and acceleration, extracts candidate keyframes, and refines them under weighted position, rotation, and gripper costs.
The extracted trajectory indices are then mapped to video timestamps and frame numbers, producing candidate annotation anchors in the visual stream.
These automatically extracted keyframes are not treated as final progress labels; annotators can add, remove, or modify points according to task semantics.

\subsection{Curated Split and Data Usage}
\label{app:dataset_split}

After constructing the full annotation inventory, we curate a train--evaluation dataset for supervised training and diagnostic evaluation.
The split is designed to evaluate two complementary forms of generalization: semantic task familiarity and progress-pattern recognition.
Semantic task familiarity is measured by separating evaluation records into seen and unseen task units, while progress-pattern recognition is measured by separating expert-like trajectories from trajectories that contain non-monotonic or regressive progress patterns.

Seen and unseen are defined at the semantic task-unit level with respect to the annotation-derived instruction data used for supervised training.
A seen evaluation record belongs to a task unit that appears in the training split, but the evaluated video record and its progress annotations are held out.
An unseen evaluation record belongs to a task unit excluded from training instruction construction.
Progress-pattern type is defined by the annotated progress trajectory: expert records correspond to trajectories without annotated progress regression, while non-expert records contain at least one regressive progress transition.
Combining the semantic and progress-pattern axes yields four held-out evaluation buckets: expert seen, expert unseen, non-expert seen, and non-expert unseen.

To prevent leakage between training and evaluation, all progress annotations in the held-out split are excluded from prompt generation, data augmentation, in-context extension, and model fine-tuning.
Records from the same physical execution, including different camera views of the same episode, are assigned to the same split.
Within each evaluation bucket, each record corresponds to a distinct task unit, which prevents the benchmark from being inflated by repeated records of the same semantic task within a bucket.
The resulting curated split is summarized in Table~\ref{tab:split_summary}.

\begin{table*}[t]
\centering
\small
\setlength{\tabcolsep}{3pt}
\renewcommand{\arraystretch}{1.12}
\begin{tabularx}{\linewidth}{@{}
  >{\raggedright\arraybackslash}p{0.19\linewidth}
  >{\raggedleft\arraybackslash}p{0.105\linewidth}
  >{\raggedleft\arraybackslash}p{0.09\linewidth}
  >{\raggedleft\arraybackslash}p{0.12\linewidth}
  >{\centering\arraybackslash}p{0.15\linewidth}
  >{\centering\arraybackslash}X
@{}}
\toprule
Split & Records & Tasks & Points
& \multicolumn{1}{c}{Expert / Non-exp.}
& \multicolumn{1}{c}{ARX / DROID / Sim.} \\
\midrule
Train
& 24,652 & 13,996 & 331,762
& 15,467 / 9,185
& 10,357 / 13,457 / 838 \\

Expert seen
& 1,043 & 1,043 & 11,769
& 1,043 / 0
& 356 / 641 / 46 \\

Expert unseen
& 1,043 & 1,043 & 12,333
& 1,043 / 0
& 356 / 641 / 46 \\

Non-expert seen
& 713 & 713 & 9,458
& 0 / 713
& 540 / 165 / 8 \\

Non-expert unseen
& 716 & 716 & 9,850
& 0 / 716
& 540 / 168 / 8 \\

\midrule
Curated total
& 28,167 & 15,206 & 375,172
& 17,553 / 10,614
& 12,149 / 15,072 / 946 \\
\bottomrule
\end{tabularx}
\caption{
Training and held-out evaluation split summary for the curated dataset.
Records are video-level annotation units, and tasks denote semantic
manipulation tasks. The source column reports ARX-data,
DROID-derived, and simulated records, where simulated records combine LIBERO
and VLABench.
}
\label{tab:split_summary}
\end{table*}

\subsection{\vlac{} Training Details}
\label{app:vlac_training_details}

The annotation-derived instruction data are complemented by targeted augmentations.
Robotic executions are often non-monotonic: the robot may lose contact, move the object away from the goal, undo partial success, or recover after failure.
To expose the model to such cases, we construct counterfactual reverse-progress examples from pairs of annotated progress points.
An augmented clip traverses the visual segment from a later state back to an earlier state and then returns forward, while the target progress curve decreases during the reversed portion and increases during the recovery portion.
This creates explicit regression-and-recovery supervision without additional robot rollouts and discourages the shortcut that later frames are always more complete.

When object boxes and gripper keypoints are available at consecutive progress points, we generate grounding-based rationales as additional training targets.
The procedure compares object centers, gripper positions, and task-relevant distances across time, then verbalizes geometric evidence that supports increasing, stagnant, or decreasing progress.
We retain rationales only when the geometric evidence is consistent with the annotated progress direction, so this component aligns numerical progress prediction with visible spatial evidence rather than introducing a separate model module.

For episodes with robot trajectory information, we derive action-conditioned samples from kinematic keyframes or adjacent progress points.
The input may contain one to four selected camera views, wrist views when available, the task instruction, the current robot state, and an optional action-intent sentence.
The target is a relative action delta between the start and end states of the interval, expressed in millimeters for translation, degrees for rotation, and percentage units for gripper motion.
Intervals with negligible motion are filtered out.
This supervision helps the critic relate progress diagnosis and correction plans to executable robot motion, while the benchmark and downstream use still evaluate \vlac\ as a video-language critic rather than a policy.

The supervised fine-tuning corpus is centered on annotation-derived robot data and complemented with public multimodal, robotics, progress-reasoning, and spatial/video reasoning datasets.
The annotation-derived component supplies the process supervision needed by \vlac: task planning, progress estimation, state/action description, progress explanation, failure diagnosis, correction planning, grounding-aware reasoning, and action-conditioned prediction.
It also contains both expert-like executions and non-expert trajectories with stagnation, regression, failed attempts, and recovery behavior, which is essential for training a critic rather than a final-success classifier.
Auxiliary datasets are included to preserve broad visual instruction following and robot-centered reasoning capabilities.
Table~\ref{tab:training_data_mixture} summarizes the mixture.

\begin{table}[t]
\centering
\small
\setlength{\tabcolsep}{4pt}
\renewcommand{\arraystretch}{1.15}
\begin{tabularx}{\linewidth}{
  >{\raggedright\arraybackslash}p{0.26\linewidth}
  >{\raggedright\arraybackslash}X
  >{\raggedright\arraybackslash}p{0.20\linewidth}
}
\toprule
Data group & Role in training & Used amount \\
\midrule
Annotation-derived robot data
& Robotic process understanding, failure diagnosis, correction, and action-aware reasoning
& 4.76M selected samples \\
LLaVA-style instruction data~\citep{liu2023visual}
& General visual instruction following
& 200K converted samples \\
RoboReward~\citep{lee2026roboreward}
& Robot reward and trajectory-quality judgment
& 45K examples \\
RoboVQA~\citep{sermanet2024robovqa}
& Robot-centered VQA and long-horizon reasoning
& 50K samples \\
Spatial QA data~\citep{xie2026spatialqa}
& Spatial relation and logical reasoning
& 250K samples \\
ProgressLM CoT~\citep{zhang2026progresslmprogressreasoningvisionlanguage}
& Progress reasoning from partial observations
& 24K CoT SFT samples \\
MMSI-Video-Bench~\citep{lin2025mmsi}
& Video-based spatial intelligence
& 0.7K samples  \\
\bottomrule
\end{tabularx}
\caption{Training data mixture used for supervised fine-tuning. Annotation-derived robot data form the core of the corpus, while auxiliary public datasets preserve general multimodal reasoning, robotics reasoning, progress estimation, and spatial/video understanding.}
\label{tab:training_data_mixture}
\end{table}

All components are converted into the same conversation format.
This unified format allows image, video, language, grounding, and action-conditioned examples to be optimized under a single supervised fine-tuning interface.
\vlac\ uses Qwen3-VL-30B-A3B-Instruct as the base backbone and is fine-tuned as a multimodal instruction-following progress critic.
Training samples are formatted as multimodal conversations containing task instructions, task plans, images or video frames, and optional robot/action context.
The model is trained to generate structured textual responses that include progress estimates, state and action descriptions, progress explanations, success/failure analysis, correction plans, grounding labels, and relative action deltas.
We use supervised fine-tuning with an autoregressive language-modeling objective over assistant responses, so all prediction and explanation tasks share a unified generative interface.
During training, we use an effective global batch size of 2304, a maximum sequence length of 15,240 tokens, and a peak learning rate of $3\times10^{-5}$ with cosine decay.
Training is conducted with MS-SWIFT and DeepSpeed ZeRO-3 on 144 NVIDIA A800-SXM4-80GB GPUs, and the full run takes 16 days.

\subsection{VPB Task Definition and Metrics}
\label{app:vpb_details}

VPB evaluates video-language progress prediction over a dataset
\(\mathcal{D} = \{(l_i, \mathbf{o}_i, \mathbf{P}_i^{\text{key}}, \mathbf{p}_i)\}_{i=1}^{N}\),
where \(l_i\) is the natural language instruction describing the task,
\(\mathbf{o}_i = (o_{i,1}, o_{i,2}, \ldots, o_{i,T_i})\) is the observation video consisting of \(T_i\) frames,
\(\mathbf{P}_i^{\text{key}} = \{(t_{i,m}, p_{i,m}^{\text{key}})\}_{m=1}^{K_i}\) is a set of \(K_i\) keyframe annotations, where
\(t_{i,m} \in \{1, \ldots, T_i\}\) denotes a frame index and
\(p_{i,m}^{\text{key}} \in [-100, 100]\) is the corresponding progress value,
and \(\mathbf{p}_i = (p_{i,1}, \ldots, p_{i,T_i})\) is the dense ground-truth progress trajectory.
The progress value \(p \in [-100, 100]\) represents the task completion status:
\(p = 100\) indicates full completion, \(p = 0\) denotes the initial state, and
\(p < 0\) represents regressive states worse than the initial setup.
Given sparse keyframe annotations \(\mathbf{P}_i^{\text{key}}\) with
\(t_{i,1} < t_{i,2} < \cdots < t_{i,K_i}\), the dense trajectory
\(\mathbf{p}_i\) is obtained via linear interpolation between adjacent keyframes:
\begin{equation}
p_{i,t} = p_{i,m}^{\text{key}} +
\frac{t - t_{i,m}}{t_{i,m+1} - t_{i,m}}
\left(p_{i,m+1}^{\text{key}} - p_{i,m}^{\text{key}}\right),
\quad
t_{i,m} \leq t \leq t_{i,m+1}, \,
m \in \{1, \ldots, K_i-1\}.
\end{equation}
Given an instruction \(l_i\) and video \(\mathbf{o}_i\), a method should output a progress sequence
\(\hat{\mathbf{p}}_i = (\hat{p}_{i,1}, \ldots, \hat{p}_{i,T_i})\).
The predicted trajectory \(\hat{\mathbf{p}}_i\) is evaluated against the ground-truth \(\mathbf{p}_i\) across multiple metrics.

VPB evaluates progress understanding at three complementary scopes: global trajectory, terminal state, and local transitions. To accommodate fine-grained analysis across diverse execution regimes (e.g., seen vs. unseen semantics, expert vs. regressive patterns), all metrics are formally defined with respect to an arbitrary evaluation subset $\mathcal{S} \subseteq \mathcal{D}$.

\paragraph{Global-Level Progress Metrics}
\label{sec:vpb_global_metrics}
Global-level evaluation measures whether a model recovers the continuous progress trajectory of the entire video. For a given record $i$, we first compute the record-level metrics and then report their expectation over the subset $\mathcal{S}$.

\subparagraph{Progress Rank Correlation.}
Progress Rank Correlation (PRC) measures whether the predicted sequence preserves the temporal ordering of the annotated progress states using Spearman rank correlation $\rho_S$:
\begin{equation}
\mathrm{PRC}_i = \rho_S \left( \hat{\mathbf{p}}_i, \mathbf{p}_i \right),
\qquad
\mathrm{PRC}(\mathcal{S}) = \frac{1}{|\mathcal{S}|} \sum_{i\in\mathcal{S}} \mathrm{PRC}_i.
\end{equation}
Higher PRC indicates better agreement between the predicted and ground-truth progress orderings.

\subparagraph{Value-Order Correlation.}
Value-Order Correlation (VOC) follows the rank-correlation evaluation introduced by GVL for value prediction on expert videos~\citep{ma2024vision}. It assesses whether the predicted values increase monotonically with the chronological progression of frames. Since chronological monotonicity is a structural property unique to flawless demonstrations, VOC is evaluated exclusively on expert data. For an applicable subset $\mathcal{S}$ consisting of expert executions, the record-level and set-level VOC are defined as:
\begin{equation}
\mathrm{VOC}_i = \rho_S \left( \hat{\mathbf{p}}_i, (1,2,\ldots,T_i) \right),
\qquad
\mathrm{VOC}(\mathcal{S}) = \frac{1}{|\mathcal{S}|} \sum_{i\in\mathcal{S}} \mathrm{VOC}_i.
\end{equation}

\subparagraph{Mean Absolute Error.}
Mean Absolute Error (MAE) evaluates the absolute calibration error of the predictions mapped to the VPB scale:
\begin{equation}
\mathrm{MAE}_i = \frac{1}{T_i} \sum_{t=1}^{T_i} \left| \hat{p}_{i,t}-p_{i,t} \right|,
\qquad
\mathrm{MAE}(\mathcal{S}) = \frac{1}{|\mathcal{S}|} \sum_{i\in\mathcal{S}} \mathrm{MAE}_i.
\end{equation}

\paragraph{Terminal-State Recognition Metrics}
\label{sec:vpb_terminal_metrics}
Terminal-state metrics evaluate whether a method reliably discriminates the ultimate task outcome. We use $90\%$ as a near-completion threshold to tolerate minor annotation and prediction noise. A record is deemed terminally successful if its final progress achieves a predefined completion threshold ($\ge 90$):
\begin{equation}
y_i^T = \mathbb{I} \left[ p_{i,T_i}\geq 90 \right],
\qquad
\hat{y}_i^T = \mathbb{I} \left[ \hat{p}_{i,T_i}\geq 90 \right].
\end{equation}
Aggregating these binary predictions over the subset $\mathcal{S}$ yields the total counts of true positives ($\mathrm{TP}$), false positives ($\mathrm{FP}$), true negatives ($\mathrm{TN}$), and false negatives ($\mathrm{FN}$). We report the Terminal-State Accuracy ($\mathrm{TSA}$) as the overall classification accuracy across $\mathcal{S}$:
\begin{equation}
\mathrm{TSA}(\mathcal{S}) = \frac{\mathrm{TP}+\mathrm{TN}}{\mathrm{TP}+\mathrm{FP}+\mathrm{TN}+\mathrm{FN}}.
\end{equation}
To account for class imbalance within specific evaluation buckets, we additionally compute the class-wise F1 scores for successful states ($\mathrm{F1}_{S}$) and failed/incomplete states ($\mathrm{F1}_{F}$):
\begin{equation}
\mathrm{F1}_{S}(\mathcal{S}) = \frac{2\mathrm{TP}}{2\mathrm{TP}+\mathrm{FP}+\mathrm{FN}},
\qquad
\mathrm{F1}_{F}(\mathcal{S}) = \frac{2\mathrm{TN}}{2\mathrm{TN}+\mathrm{FP}+\mathrm{FN}},
\end{equation}
and report their macro average:
\begin{equation}
\mathrm{MacroF1}_{T}(\mathcal{S}) = \frac{1}{2} \left( \mathrm{F1}_{S}(\mathcal{S}) + \mathrm{F1}_{F}(\mathcal{S}) \right).
\end{equation}

\paragraph{Local Progress Direction Metrics}
\label{sec:vpb_direction_metrics}

Local progress-direction metrics are computed on adjacent semantic
anchors and evaluate whether a model assigns larger directional scores
to event-level progress changes of the corresponding type. For each
trajectory \(i\), let
\[
\mathbf{P}^{\mathrm{key}}_i
=
\{(t_{i,m},p^{\mathrm{key}}_{i,m})\}_{m=1}^{K_i}
\]
denote the annotated semantic-anchor progress values. Predictions are
sampled at the same anchor frames, yielding
\[
\hat{\mathbf{P}}^{\mathrm{key}}_i
=
\{(t_{i,m},\hat p^{\mathrm{key}}_{i,m})\}_{m=1}^{K_i}.
\]
When a method does not produce a prediction exactly at an anchor frame,
\(\hat p^{\mathrm{key}}_{i,m}\) is obtained by linear interpolation over
the method's sampled prediction sequence.

For each adjacent anchor pair \((m,m+1)\), define the annotated and
predicted local progress differences as
\begin{equation}
\Delta p^{\mathrm{key}}_{i,m}
=
p^{\mathrm{key}}_{i,m+1}
-
p^{\mathrm{key}}_{i,m},
\qquad
\Delta \hat p^{\mathrm{key}}_{i,m}
=
\hat p^{\mathrm{key}}_{i,m+1}
-
\hat p^{\mathrm{key}}_{i,m}.
\end{equation}
For an evaluation subset \(\mathcal{S}\), let
\begin{equation}
\mathcal{M}(\mathcal{S})
=
\{(i,m): i\in\mathcal{S},\;1\le m<K_i\}
\end{equation}
be the set of evaluated semantic-anchor transitions.

We formulate local direction evaluation as two one-vs-rest ranking
problems. The improvement target and score are
\begin{equation}
y^{+}_{i,m}
=
\mathbb{I}
\left[
\Delta p^{\mathrm{key}}_{i,m}>0
\right],
\qquad
s^{+}_{i,m}
=
\Delta \hat p^{\mathrm{key}}_{i,m}.
\end{equation}
The regression target and score are
\begin{equation}
y^{-}_{i,m}
=
\mathbb{I}
\left[
\Delta p^{\mathrm{key}}_{i,m}<0
\right],
\qquad
s^{-}_{i,m}
=
-\Delta \hat p^{\mathrm{key}}_{i,m}.
\end{equation}
Thus, larger \(s^{+}_{i,m}\) ranks a transition as stronger predicted
local advancement, while larger \(s^{-}_{i,m}\) ranks it as stronger
predicted local reversal.

The improvement and regression average precision scores are
\begin{equation}
\mathrm{AP}_{+}(\mathcal{S})
=
\mathrm{AP}
\left(
\{y^{+}_{i,m}\}_{(i,m)\in\mathcal{M}(\mathcal{S})},
\{s^{+}_{i,m}\}_{(i,m)\in\mathcal{M}(\mathcal{S})}
\right),
\end{equation}
\begin{equation}
\mathrm{AP}_{-}(\mathcal{S})
=
\mathrm{AP}
\left(
\{y^{-}_{i,m}\}_{(i,m)\in\mathcal{M}(\mathcal{S})},
\{s^{-}_{i,m}\}_{(i,m)\in\mathcal{M}(\mathcal{S})}
\right).
\end{equation}
Both scores are computed over the same transition set
\(\mathcal{M}(\mathcal{S})\). The local direction macro AP is
\begin{equation}
\mathrm{MacroAP}_{D}(\mathcal{S})
=
\frac{1}{2}
\left(
\mathrm{AP}_{+}(\mathcal{S})
+
\mathrm{AP}_{-}(\mathcal{S})
\right).
\end{equation}

For a binary ranking problem
\(\{(y_j,s_j)\}_{j=1}^{M}\) with \(N_{+}=\sum_{j=1}^{M}y_j\) target
instances, let \(\pi\) be the permutation that sorts examples by
descending score. Average precision is computed as
\begin{equation}
\mathrm{AP}
=
\frac{1}{N_{+}}
\sum_{r=1}^{M}
y_{\pi_r}
\frac{\sum_{q=1}^{r} y_{\pi_q}}{r}.
\end{equation}

%% file: sections/appendix_teleop.tex
\section{Teleoperation Assistance}
\label{app:teleop_assistance}

VR teleoperation provides an intuitive interface for collecting robot demonstrations, but packet loss, network jitter, packet reordering, and deadline misses may leave individual control slots without a valid command. A missing slot does not imply zero-motion intent, while holding the robot pose, repeating the last command, or applying temporal interpolation can introduce pauses or stale motion and cannot reliably recover task-dependent human intent.

\begin{figure}[htbp]
    \centering
    \includegraphics[width=0.98\linewidth]{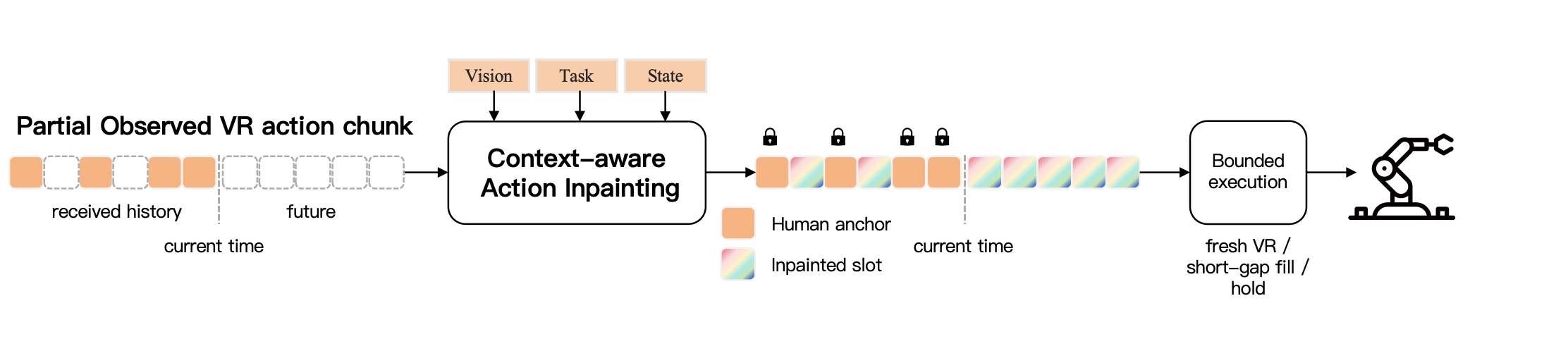}
    \caption{Overview of the VLA-based teleoperation assistance module. Observed VR commands are preserved as human anchors, while briefly missing control slots are inpainted using the deployed VLA policy and executed through a gated controller.}
    \label{fig:action_inpainting}
\end{figure}

As illustrated in Fig.~\ref{fig:action_inpainting}, we address this problem through \emph{context-aware human-intent inpainting}. Importantly, the inpainting model is not a separate teleoperation policy; it reuses the same VLA policy $\pi_\theta$ currently deployed on the robots for autonomous execution and continuously updated during post-training. We regard the VR command stream as a partially observed action trajectory. Incoming commands are assigned to fixed-rate control slots, forming a partially observed action chunk $\mathbf{A}^{\mathrm{obs}}_t$ and an availability mask $\mathbf{m}_t$. The deployed VLA completes the missing slots according to
\begin{equation}
    \hat{\mathbf{A}}_t
    \sim
    p_{\theta}\!\left(
        \mathbf{A}
        \mid
        \mathbf{o}_t,\mathbf{s}_t,\ell,
        \mathbf{A}^{\mathrm{obs}}_t,\mathbf{m}_t
    \right),
    \qquad
    \hat{\mathbf{A}}_{t,i}
    =
    \mathbf{A}^{\mathrm{obs}}_{t,i}
    \quad
    \forall i \text{ such that } m_{t,i}=1,
\end{equation}
where $p_\theta$ is the conditional action distribution induced by the deployed VLA policy $\pi_\theta$, and $\mathbf{o}_t$, $\mathbf{s}_t$, and $\ell$ denote the visual observation, robot state, and task instruction.

At inference time, unknown slots are generated jointly as a short-horizon, context-consistent action sequence, while every observed slot is preserved through hard projection. A fresh VR command is always executed directly; the inpainted action is used only when the current slot is briefly unavailable. A signal-quality gate disables model-generated motion when the latest command becomes stale or the interruption exceeds a predefined duration, causing the robot to hold its current pose until direct human control resumes.

%% file: refs.bib
@string{neurips = "Advances in Neural Information Processing Systems (NeurIPS)"}

@string{icra = "IEEE International Conference on Robotics and Automation (ICRA)"}

@string{rss = "Robotics: Science and Systems Conference (RSS)"}

@string{chi = "ACM Conference on Human Factors in Computing Systems (CHI)"}

@article{alakuijala2024videolanguagecritic,
  title   = {Video-Language Critic: Transferable Reward Functions for Language-Conditioned Robotics},
  author  = {Alakuijala, Minttu and McLean, Reginald and Woungang, Isaac and Farsad, Nariman and Kaski, Samuel and Marttinen, Pekka and Yuan, Kai},
  journal = {arXiv preprint arXiv:2405.19988},
  year    = {2024},
  url     = {https://arxiv.org/abs/2405.19988}
}

@inproceedings{andrychowicz2017hindsight,
  title={Hindsight experience replay},
  author={Andrychowicz, Marcin and Wolski, Filip and Ray, Alex and Schneider, Jonas and Fong, Rachel and Welinder, Peter and McGrew, Bob and Tobin, Josh and Abbeel, Pieter and Zaremba, Wojciech},
  booktitle={NeurIPS},
  year={2017}
}

@article{black2024pi_0,
  title={$pi\_0 $: A Vision-Language-Action Flow Model for General Robot Control},
  author={Black, Kevin and Brown, Noah and Driess, Danny and Esmail, Adnan and Equi, Michael and Finn, Chelsea and Fusai, Niccolo and Groom, Lachy and Hausman, Karol and Ichter, Brian and others},
  journal={arXiv preprint arXiv:2410.24164},
  year={2024}
}

@article{budzianowski2025opengvl,
  title={OpenGVL--Benchmarking Visual Temporal Progress for Data Curation},
  author={Budzianowski, Pawe{\l} and Wi{\'s}nios, Emilia and G{\'o}ral, Gracjan and Kulakov, Igor and Petrenko, Viktor and Walas, Krzysztof},
  journal={arXiv preprint arXiv:2509.17321},
  year={2025}
}

@article{chen2021learning,
  title={Learning Generalizable Robotic Reward Functions from" In-The-Wild" Human Videos},
  author={Chen, Annie S and Nair, Suraj and Finn, Chelsea},
  journal={RSS},
  year={2021}
}

@article{chen2025conrft,
  title={Conrft: A reinforced fine-tuning method for vla models via consistency policy},
  author={Chen, Yuhui and Tian, Shuai and Liu, Shugao and Zhou, Yingting and Li, Haoran and Zhao, Dongbin},
  journal={arXiv preprint arXiv:2502.05450},
  year={2025}
}

@article{chen2025sarm,
  title={SARM: Stage-Aware Reward Modeling for Long Horizon Robot Manipulation},
  author={Chen, Qianzhong and Yu, Justin and Schwager, Mac and Abbeel, Pieter and Shentu, Yide and Wu, Philipp},
  journal={arXiv preprint arXiv:2509.25358},
  year={2025}
}

@article{chen2026topreward,
  title        = {TOPReward: Token Probabilities as Hidden Zero-Shot Rewards for Robotics},
  author       = {Chen, Shirui and Harrison, Cole and Lee, Ying-Chun and Yang, Angela Jin and Ren, Zhongzheng and Ratliff, Lillian J. and Duan, Jiafei and Fox, Dieter and Krishna, Ranjay},
  journal      = {arXiv preprint arXiv:2602.19313},
  year         = {2026}
}

@inproceedings{cui2025grove,
  title={Grove: A generalized reward for learning open-vocabulary physical skill},
  author={Cui, Jieming and Liu, Tengyu and Meng, Ziyu and Yu, Jiale and Song, Ran and Zhang, Wei and Zhu, Yixin and Huang, Siyuan},
  booktitle={Proceedings of the Computer Vision and Pattern Recognition Conference},
  pages={15781--15790},
  year={2025}
}

@misc{deepseekai2025deepseekr1incentivizingreasoningcapability,
      title={DeepSeek-R1: Incentivizing Reasoning Capability in LLMs via Reinforcement Learning}, 
      author={DeepSeek-AI and Daya Guo and Dejian Yang and Haowei Zhang and Junxiao Song and Ruoyu Zhang and Runxin Xu and Qihao Zhu and Shirong Ma and Peiyi Wang and Xiao Bi and Xiaokang Zhang and Xingkai Yu and Yu Wu and Z. F. Wu and Zhibin Gou and Zhihong Shao and Zhuoshu Li and Ziyi Gao and Aixin Liu and Bing Xue and Bingxuan Wang and Bochao Wu and Bei Feng and Chengda Lu and Chenggang Zhao and Chengqi Deng and Chenyu Zhang and Chong Ruan and Damai Dai and Deli Chen and Dongjie Ji and Erhang Li and Fangyun Lin and Fucong Dai and Fuli Luo and Guangbo Hao and Guanting Chen and Guowei Li and H. Zhang and Han Bao and Hanwei Xu and Haocheng Wang and Honghui Ding and Huajian Xin and Huazuo Gao and Hui Qu and Hui Li and Jianzhong Guo and Jiashi Li and Jiawei Wang and Jingchang Chen and Jingyang Yuan and Junjie Qiu and Junlong Li and J. L. Cai and Jiaqi Ni and Jian Liang and Jin Chen and Kai Dong and Kai Hu and Kaige Gao and Kang Guan and Kexin Huang and Kuai Yu and Lean Wang and Lecong Zhang and Liang Zhao and Litong Wang and Liyue Zhang and Lei Xu and Leyi Xia and Mingchuan Zhang and Minghua Zhang and Minghui Tang and Meng Li and Miaojun Wang and Mingming Li and Ning Tian and Panpan Huang and Peng Zhang and Qiancheng Wang and Qinyu Chen and Qiushi Du and Ruiqi Ge and Ruisong Zhang and Ruizhe Pan and Runji Wang and R. J. Chen and R. L. Jin and Ruyi Chen and Shanghao Lu and Shangyan Zhou and Shanhuang Chen and Shengfeng Ye and Shiyu Wang and Shuiping Yu and Shunfeng Zhou and Shuting Pan and S. S. Li and Shuang Zhou and Shaoqing Wu and Shengfeng Ye and Tao Yun and Tian Pei and Tianyu Sun and T. Wang and Wangding Zeng and Wanjia Zhao and Wen Liu and Wenfeng Liang and Wenjun Gao and Wenqin Yu and Wentao Zhang and W. L. Xiao and Wei An and Xiaodong Liu and Xiaohan Wang and Xiaokang Chen and Xiaotao Nie and Xin Cheng and Xin Liu and Xin Xie and Xingchao Liu and Xinyu Yang and Xinyuan Li and Xuecheng Su and Xuheng Lin and X. Q. Li and Xiangyue Jin and Xiaojin Shen and Xiaosha Chen and Xiaowen Sun and Xiaoxiang Wang and Xinnan Song and Xinyi Zhou and Xianzu Wang and Xinxia Shan and Y. K. Li and Y. Q. Wang and Y. X. Wei and Yang Zhang and Yanhong Xu and Yao Li and Yao Zhao and Yaofeng Sun and Yaohui Wang and Yi Yu and Yichao Zhang and Yifan Shi and Yiliang Xiong and Ying He and Yishi Piao and Yisong Wang and Yixuan Tan and Yiyang Ma and Yiyuan Liu and Yongqiang Guo and Yuan Ou and Yuduan Wang and Yue Gong and Yuheng Zou and Yujia He and Yunfan Xiong and Yuxiang Luo and Yuxiang You and Yuxuan Liu and Yuyang Zhou and Y. X. Zhu and Yanhong Xu and Yanping Huang and Yaohui Li and Yi Zheng and Yuchen Zhu and Yunxian Ma and Ying Tang and Yukun Zha and Yuting Yan and Z. Z. Ren and Zehui Ren and Zhangli Sha and Zhe Fu and Zhean Xu and Zhenda Xie and Zhengyan Zhang and Zhewen Hao and Zhicheng Ma and Zhigang Yan and Zhiyu Wu and Zihui Gu and Zijia Zhu and Zijun Liu and Zilin Li and Ziwei Xie and Ziyang Song and Zizheng Pan and Zhen Huang and Zhipeng Xu and Zhongyu Zhang and Zhen Zhang},
      year={2025},
      eprint={2501.12948},
      archivePrefix={arXiv},
      primaryClass={cs.CL},
      url={https://arxiv.org/abs/2501.12948}, 
}

@article{du2023vision,
  title={Vision-language models as success detectors},
  author={Du, Yuqing and Konyushkova, Ksenia and Denil, Misha and Raju, Akhil and Landon, Jessica and Hill, Felix and de Freitas, Nando and Cabi, Serkan},
  journal={arXiv preprint arXiv:2303.07280},
  year={2023}
}

@article{escontrela2023video,
  title={Video prediction models as rewards for reinforcement learning},
  author={Escontrela, Alejandro and Adeniji, Ademi and Yan, Wilson and Jain, Ajay and Peng, Xue Bin and Goldberg, Ken and Lee, Youngwoon and Hafner, Danijar and Abbeel, Pieter},
  journal={Advances in Neural Information Processing Systems},
  volume={36},
  pages={68760--68783},
  year={2023}
}

@article{ghasemipour2025self,
  title={Self-Improving Embodied Foundation Models},
  author={Ghasemipour, Seyed Kamyar Seyed and Wahid, Ayzaan and Tompson, Jonathan and Sanketi, Pannag and Mordatch, Igor},
  journal={arXiv preprint arXiv:2509.15155},
  year={2025}
}

@article{intelligence2025pi06vla,
  title   = {$\pi^{*}_{0.6}$: A VLA That Learns From Experience},
  author  = {Physical Intelligence and Ali Amin and Raichelle Aniceto and Ashwin Balakrishna and Kevin Black and Ken Conley and Grace Connors and James Darpinian and Karan Dhabalia and Jared DiCarlo and Danny Driess and Michael Equi and Adnan Esmail and Yunhao Fang and Chelsea Finn and Catherine Glossop and Thomas Godden and Ivan Goryachev and Lachy Groom and Hunter Hancock and Karol Hausman and Gashon Hussein and Brian Ichter and Szymon Jakubczak and Rowan Jen and Tim Jones and Ben Katz and Liyiming Ke and Chandra Kuchi and Marinda Lamb and Devin LeBlanc and Sergey Levine and Adrian Li-Bell and Yao Lu and Vishnu Mano and Mohith Mothukuri and Suraj Nair and Karl Pertsch and Allen Z. Ren and Charvi Sharma and Lucy Xiaoyang Shi and Laura Smith and Jost Tobias Springenberg and Kyle Stachowicz and Will Stoeckle and Alex Swerdlow and James Tanner and Marcel Torne and Quan Vuong and Anna Walling and Haohuan Wang and Blake Williams and Sukwon Yoo and Lili Yu and Ury Zhilinsky and Zhiyuan Zhou},
  journal = {arXiv:2511.14759},
  year    = {2025}
}

@article{lambert2024rewardbench,
  title        = {RewardBench: Evaluating Reward Models for Language Modeling},
  author       = {Nathan Lambert and Valentina Pyatkin and Jacob Morrison and
                  L.~J. Miranda and Bill Yuchen Lin and Khyathi Chandu and
                  Nouha Dziri and Sachin Kumar and Tom Zick and Yejin Choi and
                  Noah A. Smith and Hannaneh Hajishirzi},
  journal      = {arXiv preprint arXiv:2403.13787},
  year         = {2024},
  url          = {https://arxiv.org/abs/2403.13787}
}

@article{lee2021generalizable,
  title={Generalizable Imitation Learning from Observation via Inferring Goal Proximity},
  author={Lee, Youngwoon and Szot, Andrew and Sun, Shao-Hua and Lim, Joseph J},
  journal={Advances in Neural Information Processing Systems},
  volume={34},
  year={2021}
}

@article{lee2026roboreward,
  title={RoboReward: General-Purpose Vision-Language Reward Models for Robotics},
  author={Lee, Tony and Wagenmaker, Andrew and Pertsch, Karl and Liang, Percy and Levine, Sergey and Finn, Chelsea},
  journal={arXiv preprint arXiv:2601.00675},
  year={2026}
}

@article{li2024vlrewardbench,
  title        = {VLRewardBench: A Challenging Benchmark for Vision--Language
                  Generative Reward Models},
  author       = {Lei Li and Yuancheng Wei and Zhihui Xie and Xuqing Yang and
                  Yifan Song and Peiyi Wang and Chenxin An and Tianyu Liu and
                  Sujian Li and Bill Yuchen Lin and Lingpeng Kong and Qi Liu},
  journal      = {arXiv preprint arXiv:2411.17451},
  year         = {2024},
  url          = {https://arxiv.org/abs/2411.17451}
}

@inproceedings{lin2024navigating,
  title={Navigating noisy feedback: Enhancing reinforcement learning with error-prone language models},
  author={Lin, Muhan and Shi, Shuyang and Guo, Yue and Chalaki, Behdad and Tadiparthi, Vaishnav and Pari, Ehsan Moradi and Stepputtis, Simon and Campbell, Joseph P and Sycara, Katia P},
  booktitle={Findings of the Association for Computational Linguistics: EMNLP 2024},
  pages={16002--16014},
  year={2024}
}

@article{liang2026robometer,
  title={Robometer: Scaling general-purpose robotic reward models via trajectory comparisons},
  author={Liang, Anthony and Korkmaz, Yigit and Zhang, Jiahui and Hwang, Minyoung and Anwar, Abrar and Kaushik, Sidhant and Shah, Aditya and Huang, Alex S and Zettlemoyer, Luke and Fox, Dieter and others},
  journal={arXiv preprint arXiv:2603.02115},
  year={2026}
}

@article{liu2023visual,
  title={Visual instruction tuning},
  author={Liu, Haotian and Li, Chunyuan and Wu, Qingyang and Lee, Yong Jae},
  journal={Advances in neural information processing systems},
  volume={36},
  pages={34892--34916},
  year={2023}
}

@article{luu2025erlvlm,
  title   = {ERL-VLM: Enhancing Rating-Based Reinforcement Learning to Effectively Leverage Feedback from Large Vision–Language Models},
  author  = {Tung M. Luu and Younghwan Lee and Donghoon Lee and Sunho Kim and Min Jun Kim and Chang D. Yoo},
  journal = {arXiv preprint arXiv:2506.12822},
  year    = {2025},
  url     = {https://arxiv.org/abs/2506.12822}
}

@inproceedings{lynch2020learning,
  title={Learning latent plans from play},
  author={Lynch, Corey and Khansari, Mohi and Xiao, Ted and Kumar, Vikash and Tompson, Jonathan and Levine, Sergey and Sermanet, Pierre},
  booktitle={CoRL},
  year={2020}
}

@inproceedings{ma2022vip,
  title={VIP: Towards Universal Visual Reward and Representation via Value-Implicit Pre-Training},
  author={Ma, Yecheng Jason and Sodhani, Shagun and Jayaraman, Dinesh and Bastani, Osbert and Kumar, Vikash and Zhang, Amy},
  booktitle={The Eleventh International Conference on Learning Representations},
  year={2022}
}

@article{ma2023eureka,
  title={Eureka: Human-level reward design via coding large language models},
  author={Ma, Yecheng Jason and Liang, William and Wang, Guanzhi and Huang, De-An and Bastani, Osbert and Jayaraman, Dinesh and Zhu, Yuke and Fan, Linxi and Anandkumar, Anima},
  journal={arXiv preprint arXiv:2310.12931},
  year={2023}
}

@inproceedings{ma2023liv,
  title={Liv: Language-image representations and rewards for robotic control},
  author={Ma, Yecheng Jason and Kumar, Vikash and Zhang, Amy and Bastani, Osbert and Jayaraman, Dinesh},
  booktitle={International Conference on Machine Learning},
  pages={23301--23320},
  year={2023},
  organization={PMLR}
}

@inproceedings{ma2024vision,
  title={Vision language models are in-context value learners},
  author={Ma, Yecheng Jason and Hejna, Joey and Fu, Chuyuan and Shah, Dhruv and Liang, Jacky and Xu, Zhuo and Kirmani, Sean and Xu, Peng and Driess, Danny and Xiao, Ted and others},
  booktitle={The Thirteenth International Conference on Learning Representations},
  year={2024}
}

@article{malik2025rewardbench2,
  title        = {RewardBench 2: Advancing Reward Model Evaluation},
  author       = {Saumya Malik and Valentina Pyatkin and Sander Land and
                  Jacob Morrison and Noah A. Smith and Hannaneh Hajishirzi and
                  Nathan Lambert},
  journal      = {arXiv preprint arXiv:2506.01937},
  year         = {2025},
  url          = {https://arxiv.org/abs/2506.01937}
}

@inproceedings{nair2022r3m,
  title={R3m: A universal visual representation for robot manipulation},
  author={Nair, Suraj and Rajeswaran, Aravind and Kumar, Vikash and Finn, Chelsea and Gupta, Abhinav},
  booktitle={CoRL},
  year={2022}
}

@article{rocamonde2023vision,
  title={Vision-language models are zero-shot reward models for reinforcement learning},
  author={Rocamonde, Juan and Montesinos, Victoriano and Nava, Elvis and Perez, Ethan and Lindner, David},
  journal={arXiv preprint arXiv:2310.12921},
  year={2023}
}

@article{sermanet2016unsupervised,
  title={Unsupervised perceptual rewards for imitation learning},
  author={Sermanet, Pierre and Xu, Kelvin and Levine, Sergey},
  journal={arXiv preprint arXiv:1612.06699},
  year={2016}
}

@inproceedings{shao2020concept,

 title={Concept2Robot: Learning Manipulation Concepts from Instructions and Human Demonstrations},

 author={Shao, Lin and Migimatsu, Toki and Zhang, Qiang and Yang, Karen and Bohg, Jeannette},

 booktitle={Proceedings of Robotics: Science and Systems (RSS)},

 year={2020},}

@misc{shao2024deepseekmathpushinglimitsmathematical,
      title={DeepSeekMath: Pushing the Limits of Mathematical Reasoning in Open Language Models}, 
      author={Zhihong Shao and Peiyi Wang and Qihao Zhu and Runxin Xu and Junxiao Song and Xiao Bi and Haowei Zhang and Mingchuan Zhang and Y. K. Li and Y. Wu and Daya Guo},
      year={2024},
      eprint={2402.03300},
      archivePrefix={arXiv},
      primaryClass={cs.CL},
      url={https://arxiv.org/abs/2402.03300}, 
}

@article{singh2025varp,
  title   = {VARP: Reinforcement Learning from Vision–Language Model Feedback with Agent-Regularized Preferences},
  author  = {Anukriti Singh and Amisha Bhaskar and Peihong Yu and Souradip Chakraborty and Ruthwik Dasyam and Amrit Bedi and Pratap Tokekar},
  journal = {arXiv preprint arXiv:2503.13817},
  year    = {2025},
  url     = {https://arxiv.org/pdf/2503.13817}
}

@article{sontakke2024roboclip,
  title={Roboclip: One demonstration is enough to learn robot policies},
  author={Sontakke, Sumedh and Zhang, Jesse and Arnold, S{\'e}b and Pertsch, Karl and B{\i}y{\i}k, Erdem and Sadigh, Dorsa and Finn, Chelsea and Itti, Laurent},
  journal={Advances in Neural Information Processing Systems},
  volume={36},
  year={2024}
}

@article{tan2025robodopamine,
    title={Robo-Dopamine: General Process Reward Modeling for High-Precision Robotic Manipulation}, 
    author={Tan, Huajie and Chen, Sixiang and Xu, Yijie and Wang, Zixiao and Ji, Yuheng and Chi, Cheng and Lyu, Yaoxu and Zhao, Zhongxia and Chen, Xiansheng and Co, Peterson and Xie, Shaoxuan and Yao, Guocai and Wang, Pengwei and Wang, Zhongyuan and Zhang, Shanghang},
    journal={arXiv preprint arXiv:2512.23703},
    year={2025}
}

@article{venkataraman2024offlinevlm,
  title   = {Real-World Offline Reinforcement Learning from Vision Language Model Feedback},
  author  = {Sreyas Venkataraman and Yufei Wang and Ziyu Wang and Zackory Erickson and David Held},
  journal = {arXiv preprint arXiv:2411.05273},
  year    = {2024},
  url     = {https://arxiv.org/abs/2411.05273}
}

@article{wang2024rlvlmf,
  title   = {RL-VLM-F: Reinforcement Learning from Vision–Language Foundation Model Feedback},
  author  = {Yufei Wang and Zhanyi Sun and Jesse Zhang and Zhou Xian and Erdem Biyik and David Held and Zackory Erickson},
  journal = {arXiv preprint arXiv:2402.03681},
  year    = {2024},
  url     = {https://arxiv.org/abs/2402.03681}
}

@misc{
  yang2023robot,
  title={Robot Fine-Tuning Made Easy: Pre-Training Rewards and Policies for Autonomous Real-World Reinforcement Learning},
  author={Jingyun Yang and Max Sobol Mark and Brandon Vu and Archit Sharma and Jeannette Bohg and Chelsea Finn},
  year={2023},
  eprint={2310.15145},
  archivePrefix={arXiv},
  primaryClass={cs.RO}
}

@article{yasunaga2025mmrewardbench,
  title        = {Multimodal RewardBench: Holistic Evaluation of Reward Models
                  for Vision--Language Models},
  author       = {Michihiro Yasunaga and Luke Zettlemoyer and
                  Marjan Ghazvininejad},
  journal      = {arXiv preprint arXiv:2502.14191},
  year         = {2025},
  url          = {https://arxiv.org/abs/2502.14191}
}

@article{zhai2025vlac,
  title={A Vision-Language-Action-Critic Model for Robotic Real-World Reinforcement Learning},
  author={Zhai, Shaopeng and Zhang, Qi and Zhang, Tianyi and Huang, Fuxian and Zhang, Haoran and Zhou, Ming and Zhang, Shengzhe and Liu, Litao and Lin, Sixu and Pang, Jiangmiao},
  journal={arXiv preprint arXiv:2509.15937},
  year={2025}
}

@article{zhang2025rewind,
  title   = {ReWiND: Language-Guided Rewards Teach Robot Policies without New Demonstrations},
  author  = {Jiahui Zhang and Yusen Luo and Abrar Anwar and Sumedh A. Sontakke and Joseph J. Lim and Jesse Thomason and Erdem Bıyık and Jesse Zhang},
  journal = {arXiv preprint arXiv:2505.10911},
  year    = {2025},
  url     = {https://arxiv.org/abs/2505.10911}
}

@article{zhang2026progresslmprogressreasoningvisionlanguage,
      title={PROGRESSLM: Towards Progress Reasoning in Vision-Language Models}, 
      author={Jianshu Zhang and Chengxuan Qian and Haosen Sun and Haoran Lu and Dingcheng Wang and Letian Xue and Han Liu},
      year={2026},
      journal={arXiv preprint arXiv:2601.15224},
}

@article{mao2026arm,
  title={ARM: Advantage Reward Modeling for Long-Horizon Manipulation},
  author={Mao, Yiming and Yu, Zixi and Mao, Weixin and Li, Yinhao and Hu, Qirui and Lan, Zihan and Zhu, Minzhao and Chen, Hua},
  journal={arXiv preprint arXiv:2604.03037},
  year={2026}
}

@article{yu2026chi_,
  title={{$\chi_0$}: Resource-Aware Robust Manipulation via Taming Distributional Inconsistencies},
  author={Yu, Checheng and Sima, Chonghao and Jiang, Gangcheng and Zhang, Hai and Mai, Haoguang and Li, Hongyang and Wang, Huijie and Chen, Jin and Wu, Kaiyang and Chen, Li and others},
  journal={arXiv preprint arXiv:2602.09021},
  year={2026}
}

@article{krack2026rewarding,
  title={Rewarding DINO: Predicting Dense Rewards with Vision Foundation Models},
  author={Krack, Pierre and J{\"u}lg, Tobias and Burgard, Wolfram and Walter, Florian},
  journal={arXiv preprint arXiv:2603.16978},
  year={2026}
}

@article{wu2026large,
  title={Large Reward Models: Generalizable Online Robot Reward Generation with Vision-Language Models},
  author={Wu, Yanru and Yuan, Weiduo and Qi, Ang and Guizilini, Vitor and Mao, Jiageng and Wang, Yue},
  journal={arXiv preprint arXiv:2603.16065},
  year={2026}
}

@article{ye2025robofac,
  title={Robofac: A comprehensive framework for robotic failure analysis and correction},
  author={Ye, Zewei and Lu, Weifeng and Ye, Minghao and Lin, Tao and Yang, Shuo and Yan, Junchi and Zhao, Bo},
  journal={arXiv preprint arXiv:2505.12224},
  year={2025}
}

@article{khazatsky2024droid,
  title={Droid: A large-scale in-the-wild robot manipulation dataset},
  author={Khazatsky, Alexander and Pertsch, Karl and Nair, Suraj and Balakrishna, Ashwin and Dasari, Sudeep and Karamcheti, Siddharth and Nasiriany, Soroush and Srirama, Mohan Kumar and Chen, Lawrence Yunliang and Ellis, Kirsty and others},
  journal={arXiv preprint arXiv:2403.12945},
  year={2024}
}

@article{liu2023libero,
  title={Libero: Benchmarking knowledge transfer for lifelong robot learning},
  author={Liu, Bo and Zhu, Yifeng and Gao, Chongkai and Feng, Yihao and Liu, Qiang and Zhu, Yuke and Stone, Peter},
  journal={Advances in Neural Information Processing Systems},
  volume={36},
  pages={44776--44791},
  year={2023}
}

@inproceedings{zhang2025vlabench,
  title={Vlabench: A large-scale benchmark for language-conditioned robotics manipulation with long-horizon reasoning tasks},
  author={Zhang, Shiduo and Xu, Zhe and Liu, Peiju and Yu, Xiaopeng and Li, Yuan and Gao, Qinghui and Fei, Zhaoye and Yin, Zhangyue and Wu, Zuxuan and Jiang, Yu-Gang and others},
  booktitle={Proceedings of the IEEE/CVF International Conference on Computer Vision},
  pages={11142--11152},
  year={2025}
}

@inproceedings{sermanet2024robovqa,
  title={Robovqa: Multimodal long-horizon reasoning for robotics},
  author={Sermanet, Pierre and Ding, Tianli and Zhao, Jeffrey and Xia, Fei and Dwibedi, Debidatta and Gopalakrishnan, Keerthana and Chan, Christine and Dulac-Arnold, Gabriel and Maddineni, Sharath and Joshi, Nikhil J and others},
  booktitle={2024 IEEE International Conference on Robotics and Automation (ICRA)},
  pages={645--652},
  year={2024},
  organization={IEEE}
}

@article{xie2026spatialqa,
  title={Spatialqa: A benchmark for evaluating spatial logical reasoning in vision-language models},
  author={Xie, Yuechen and Zhang, Xiaoyan and Shan, Yicheng and Zhu, Hao and Tang, Rui and Wei, Rong and Song, Mingli and Wan, Yuanyu and Song, Jie},
  journal={arXiv preprint arXiv:2602.20901},
  year={2026}
}

@article{lin2025mmsi,
  title={MMSI-Video-Bench: A Holistic Benchmark for Video-Based Spatial Intelligence},
  author={Lin, Jingli and Xu, Runsen and Zhu, Shaohao and Yang, Sihan and Cao, Peizhou and Ran, Yunlong and Hu, Miao and Zhu, Chenming and Xie, Yiman and Long, Yilin and others},
  journal={arXiv preprint arXiv:2512.10863},
  year={2025}
}

@inproceedings{smith2024grow,
  title={Grow your limits: Continuous improvement with real-world rl for robotic locomotion},
  author={Smith, Laura and Cao, Yunhao and Levine, Sergey},
  booktitle={2024 IEEE International Conference on Robotics and Automation (ICRA)},
  pages={10829--10836},
  year={2024},
  organization={IEEE}
}

@article{smith2023demonstrating,
  title={Demonstrating a walk in the park: Learning to walk in 20 minutes with model-free reinforcement learning},
  author={Smith, Laura and Kostrikov, Ilya and Levine, Sergey},
  journal={Robotics: Science and Systems (RSS) Demo},
  volume={2},
  number={3},
  pages={4},
  year={2023}
}

@InProceedings{pmlr-v229-hu23a,
  title = 	 {REBOOT: Reuse Data for Bootstrapping Efficient Real-World Dexterous Manipulation},
  author =       {Hu, Zheyuan and Rovinsky, Aaron and Luo, Jianlan and Kumar, Vikash and Gupta, Abhishek and Levine, Sergey},
  booktitle = 	 {Proceedings of The 7th Conference on Robot Learning},
  pages = 	 {1930--1949},
  year = 	 {2023},
  editor = 	 {Tan, Jie and Toussaint, Marc and Darvish, Kourosh},
  volume = 	 {229},
  series = 	 {Proceedings of Machine Learning Research},
  month = 	 {06--09 Nov},
  publisher =    {PMLR},
  url = 	 {https://proceedings.mlr.press/v229/hu23a.html}
}

@inproceedings{kumar2024practice,
    title={Practice Makes Perfect: Planning to Learn Skill Parameter Policies}, 
    author={Nishanth Kumar and Tom Silver and Willie McClinton and Linfeng Zhao and Stephen Proulx and Tomás Lozano-Pérez and Leslie Pack Kaelbling and Jennifer Barry},
    year={2024},
    booktitle={Robotics: Science and Systems (RSS)}
}

@article{zhou2024autonomous,
    title={Autonomous Improvement of Instruction Following Skills via Foundation Models},
    author={Zhiyuan Zhou and Pranav Atreya and Abraham Lee and Homer Walke and Oier Mees and Sergey Levine},
    journal = {arXiv preprint arXiv:407.20635},
    year={2024},
}

@misc{luo2024hilserl,
      title={Precise and Dexterous Robotic Manipulation via Human-in-the-Loop Reinforcement Learning},
      author={Jianlan Luo and Charles Xu and Jeffrey Wu and Sergey Levine},
      year={2024},
      eprint={2410.21845},
      archivePrefix={arXiv},
      primaryClass={cs.RO}
}

@article{li2025reinforcement,
  title={Reinforcement Learning with Action Chunking},
  author={Li, Qiyang and Zhou, Zhiyuan and Levine, Sergey},
  journal={arXiv preprint arXiv:2507.07969},
  year={2025}
}

@article{kang2025forget,
  title={A Forget-and-Grow Strategy for Deep Reinforcement Learning Scaling in Continuous Control},
  author={Kang, Zilin and Hu, Chenyuan and Luo, Yu and Yuan, Zhecheng and Zheng, Ruijie and Xu, Huazhe},
  journal={arXiv preprint arXiv:2507.02712},
  year={2025}
}

@article{zhou2024efficient,
  title={Efficient online reinforcement learning fine-tuning need not retain offline data},
  author={Zhou, Zhiyuan and Peng, Andy and Li, Qiyang and Levine, Sergey and Kumar, Aviral},
  journal={arXiv preprint arXiv:2412.07762},
  year={2024}
}

@article{lv2025flow,
  title={Flow-Based Policy for Online Reinforcement Learning},
  author={Lv, Lei and Li, Yunfei and Luo, Yu and Sun, Fuchun and Kong, Tao and Xu, Jiafeng and Ma, Xiao},
  journal={arXiv preprint arXiv:2506.12811},
  year={2025}
}

@article{park2025flow,
  title={Flow q-learning},
  author={Park, Seohong and Li, Qiyang and Levine, Sergey},
  journal={arXiv preprint arXiv:2502.02538},
  year={2025}
}

@article{kim24openvla,
    title={OpenVLA: An Open-Source Vision-Language-Action Model},
    author={{Moo Jin} Kim and Karl Pertsch and Siddharth Karamcheti and Ted Xiao and Ashwin Balakrishna and Suraj Nair and Rafael Rafailov and Ethan Foster and Grace Lam and Pannag Sanketi and Quan Vuong and Thomas Kollar and Benjamin Burchfiel and Russ Tedrake and Dorsa Sadigh and Sergey Levine and Percy Liang and Chelsea Finn},
    journal = {arXiv preprint arXiv:2406.09246},
    year={2024}
}

@article{pertsch2025fast,
  title={Fast: Efficient action tokenization for vision-language-action models},
  author={Pertsch, Karl and Stachowicz, Kyle and Ichter, Brian and Driess, Danny and Nair, Suraj and Vuong, Quan and Mees, Oier and Finn, Chelsea and Levine, Sergey},
  journal={arXiv preprint arXiv:2501.09747},
  year={2025}
}

@article{kim2025fine,
  title={Fine-Tuning Vision-Language-Action Models: Optimizing Speed and Success},
  author={Kim, Moo Jin and Finn, Chelsea and Liang, Percy},
  journal={arXiv preprint arXiv:2502.19645},
  year={2025}
}

@article{chi2023diffusion,
  title={Diffusion policy: Visuomotor policy learning via action diffusion},
  author={Chi, Cheng and Xu, Zhenjia and Feng, Siyuan and Cousineau, Eric and Du, Yilun and Burchfiel, Benjamin and Tedrake, Russ and Song, Shuran},
  journal={The International Journal of Robotics Research},
  pages={02783649241273668},
  year={2023},
  publisher={SAGE Publications Sage UK: London, England}
}

@article{he2024aligniql,
  title={Aligniql: Policy alignment in implicit q-learning through constrained optimization},
  author={He, Longxiang and Shen, Li and Tan, Junbo and Wang, Xueqian},
  journal={arXiv preprint arXiv:2405.18187},
  year={2024}
}

@inproceedings{dingconsistency,
  title={Consistency Models as a Rich and Efficient Policy Class for Reinforcement Learning},
  author={Ding, Zihan and Jin, Chi},
  booktitle={The Twelfth International Conference on Learning Representations},
  year={2024}
}

@misc{zhai2024buildingopenendedembodiedagent,
      title={Building Open-Ended Embodied Agent via Language-Policy Bidirectional Adaptation}, 
      author={Shaopeng Zhai and Jie Wang and Tianyi Zhang and Fuxian Huang and Qi Zhang and Ming Zhou and Jing Hou and Yu Qiao and Yu Liu},
      year={2024},
      eprint={2401.00006},
      archivePrefix={arXiv},
      primaryClass={cs.AI},
      url={https://arxiv.org/abs/2401.00006}, 
}

@misc{physicalintelligence2025pi06,
  title = {$\pi^{*}_{0.6}$: A {VLA} that Learns from Experience},
  author = {{Physical Intelligence}},
  year = {2025},
  eprint = {2511.14759},
  archivePrefix = {arXiv},
  primaryClass = {cs.RO},
  url = {https://arxiv.org/abs/2511.14759}
}

@misc{xiao2026rove,
  title = {{ROVE}: Unlocking Human Interventions for Humanoid Manipulation via Reinforcement Learning},
  author = {Xiao, Xingjian and others},
  year = {2026},
  eprint = {2606.17011},
  archivePrefix = {arXiv},
  primaryClass = {cs.RO},
  url = {https://arxiv.org/abs/2606.17011}
}

@misc{pan2026sop,
  title = {{SOP}: A Scalable Online Post-Training System for Vision-Language-Action Models},
  author = {Pan, Chao and others},
  year = {2026},
  eprint = {2601.03044},
  archivePrefix = {arXiv},
  primaryClass = {cs.RO},
  url = {https://arxiv.org/abs/2601.03044}
}

@misc{wang2026learningwhiledeploying,
  title = {Learning while Deploying: Fleet-Scale Reinforcement Learning for Generalist Robot Policies},
  author = {Wang, Yixuan and others},
  year = {2026},
  eprint = {2605.00416},
  archivePrefix = {arXiv},
  primaryClass = {cs.RO},
  url = {https://arxiv.org/abs/2605.00416}
}

@misc{xiao2025worldenv,
  title = {{World-Env}: Leveraging World Model as a Virtual Environment for {VLA} Post-Training},
  author = {Xiao, Xingjian and others},
  year = {2025},
  eprint = {2509.24948},
  archivePrefix = {arXiv},
  primaryClass = {cs.RO},
  url = {https://arxiv.org/abs/2509.24948}
}

@misc{shi2026beyond,
  title = {Beyond Imitation: Reinforcement Learning-Based Sim-Real Co-Training for Vision-Language-Action Models},
  author = {Shi, Hao and others},
  year = {2026},
  eprint = {2602.12628},
  archivePrefix = {arXiv},
  primaryClass = {cs.RO},
  url = {https://arxiv.org/abs/2602.12628}
}

@misc{sun2026atomvla,
  title = {{AtomVLA}: Scalable Post-Training for Robotic Manipulation via Predictive Latent World Models},
  author = {Sun, Xiaoyu and others},
  year = {2026},
  eprint = {2603.08519},
  archivePrefix = {arXiv},
  primaryClass = {cs.RO},
  url = {https://arxiv.org/abs/2603.08519}
}

@article{zhang2026preserving,
  title={Preserving Foundational Capabilities in Flow-Matching VLAs through Conservative SFT},
  author={Zhang, Tianyi and Zhai, Shaopeng and Zhang, Haoran and Huang, Fuxian and Zhang, Qi},
  journal={arXiv preprint arXiv:2605.08879},
  year={2026}
}

@article{intelligence2025pi_,
  title={{$\pi_{0.5}$}: A Vision-Language-Action Model with Open-World Generalization},
  author={{Physical Intelligence} and Black, Kevin and Brown, Noah and Darpinian, James and Dhabalia, Karan and Driess, Danny and Esmail, Adnan and Equi, Michael and Finn, Chelsea and Fusai, Niccolo and others},
  journal={arXiv preprint arXiv:2504.16054},
  year={2025}
}
